\documentclass[10pt,twocolumn,letterpaper]{article}

\usepackage{cvpr}              

\usepackage{times}
\usepackage{epsfig}
\usepackage{graphicx}
\usepackage{amsmath}
\usepackage{amssymb}
\usepackage{multirow}
\usepackage{cuted}
\usepackage{capt-of}
\usepackage{url}

\usepackage[pagebackref,breaklinks,colorlinks]{hyperref}

\usepackage[capitalize]{cleveref}
\crefname{section}{Sec.}{Secs.}
\Crefname{section}{Section}{Sections}
\Crefname{table}{Table}{Tables}
\crefname{table}{Tab.}{Tabs.}

\def\cvprPaperID{4328} 
\def\confName{CVPR}
\def\confYear{2022}

\begin{document}
\definecolor{YangColor}{rgb}{0.7,0,0} 
\newcommand{\yang}[1]{{\color{YangColor} \textbf{[Yang: #1]}}}
\newcommand{\revise}[1]{{#1}}
\newcommand{\rev}[1]{{#1}}

\definecolor{JimeiColor}{rgb}{0,0.6,0} 
\newcommand{\jimei}[1]{{\color{JimeiColor} \textbf{[Jimei: #1]}}}

\definecolor{VangelisColor}{rgb}{0,0,0.8} 
\newcommand{\kalo}[1]{{\color{VangelisColor} \textbf{[Vangelis: #1]}}}

\definecolor{DeepaliColor}{rgb}{0.1,0.6,0.2} 
\newcommand{\deepali}[1]{{\color{DeepaliColor} \textbf{[Deepali: #1]}}}

\definecolor{DingColor}{rgb}{1.0,0.2,0.2} 
\newcommand{\ding}[1]{{\color{DingColor} \textbf{[Ding: #1]}}}

\definecolor{JunColor}{rgb}{0.9,0.44,0.0} 
\newcommand{\jun}[1]{{\color{JunColor} \textbf{[Jun: #1]}}}

\newenvironment{my_itemize}{
\begin{description} 
  \setlength{\itemsep}{1pt}
  \setlength{\parskip}{0pt}
  \setlength{\parsep}{0pt}
  }
{\end{description}}

\newcounter{mycounter}
\newenvironment{noindlist}
 {\begin{list}{\arabic{mycounter}.~~}{\usecounter{mycounter} \labelsep=0em \labelwidth=0em \leftmargin=0em \itemindent=0em}}
 {\end{list}}
 
 \newcommand{\src}{ \text{src}}
 \newcommand{\trg}{ \text{trg}}

\newcommand{\ba}{\mathbf{a}}
\newcommand{\bb}{\mathbf{b}}
\newcommand{\bc}{\mathbf{c}}
\newcommand{\bd}{\mathbf{d}}
\newcommand{\be}{\mathbf{e}}
\newcommand{\bff}{\mathbf{f}}
\newcommand{\bg}{\mathbf{g}}
\newcommand{\bh}{\mathbf{h}}
\newcommand{\bi}{\mathbf{i}}
\newcommand{\bj}{\mathbf{j}}
\newcommand{\bk}{\mathbf{k}}
\newcommand{\bl}{\mathbf{l}}
\newcommand{\bm}{\mathbf{m}}
\newcommand{\bn}{\mathbf{n}}
\newcommand{\bo}{\mathbf{o}}
\newcommand{\bp}{\mathbf{p}}
\newcommand{\bq}{\mathbf{q}}
\newcommand{\br}{\mathbf{r}}
\newcommand{\bs}{\mathbf{s}}
\newcommand{\bt}{\mathbf{t}}
\newcommand{\bu}{\mathbf{u}}
\newcommand{\bv}{\mathbf{v}}
\newcommand{\bw}{\mathbf{w}}
\newcommand{\bx}{\mathbf{x}}
\newcommand{\by}{\mathbf{y}}
\newcommand{\bz}{\mathbf{z}}
\newcommand{\bA}{\mathbf{A}}
\newcommand{\bB}{\mathbf{B}}
\newcommand{\bC}{\mathbf{C}}
\newcommand{\bD}{\mathbf{D}}
\newcommand{\bE}{\mathbf{E}}
\newcommand{\bF}{\mathbf{F}}
\newcommand{\bG}{\mathbf{G}}
\newcommand{\bH}{\mathbf{H}}
\newcommand{\bI}{\mathbf{I}}
\newcommand{\bJ}{\mathbf{J}}
\newcommand{\bK}{\mathbf{K}}
\newcommand{\bL}{\mathbf{L}}
\newcommand{\bM}{\mathbf{M}}
\newcommand{\bN}{\mathbf{N}}
\newcommand{\bO}{\mathbf{O}}
\newcommand{\bP}{\mathbf{P}}
\newcommand{\bQ}{\mathbf{Q}}
\newcommand{\bR}{\mathbf{R}}
\newcommand{\bS}{\mathbf{S}}
\newcommand{\bT}{\mathbf{T}}
\newcommand{\bU}{\mathbf{U}}
\newcommand{\bV}{\mathbf{V}}
\newcommand{\bW}{\mathbf{W}}
\newcommand{\bX}{\mathbf{X}}
\newcommand{\bY}{\mathbf{Y}}
\newcommand{\bZ}{\mathbf{Z}}
\newcommand{\balpha}{\mbox{\boldmath$\alpha$}}
\newcommand{\bgamma}{\mbox{\boldmath$\gamma$}}
\newcommand{\bGamma}{\mbox{\boldmath$\Gamma$}}
\newcommand{\bmu}{\mbox{\boldmath$\mu$}}
\newcommand{\bphi}{\mbox{\boldmath$\phi$}}
\newcommand{\bPhi}{\mbox{\boldmath$\Phi$}}
\newcommand{\bSigma}{\mbox{\boldmath$\Sigma$}}
\newcommand{\bsigma}{\mbox{\boldmath$\sigma$}}
\newcommand{\btheta}{\mbox{\boldmath$\theta$}}

\newcommand{\mJ}{\mathcal{J}}
\newcommand{\mG}{\mathcal{G}}
\newcommand{\mE}{\mathcal{E}}
\newcommand{\mV}{\mathcal{V}}
\newcommand{\mM}{\mathcal{M}}
\newcommand{\mL}{\mathcal{L}}
\newcommand{\mU}{\mathcal{U}}
\newcommand{\mC}{\mathcal{C}}
\newcommand{\mS}{\mathcal{S}}
\newcommand{\mR}{\mathcal{R}}
\newcommand{\mD}{\mathcal{D}}
\newcommand{\mT}{\mathcal{T}}
\newcommand{\mSl}{\mathcal{S}_l}
\newcommand{\mN}{\mathcal{N}}
\newcommand{\mDll}{\mathcal{D}_{l,l'}}

\newcommand{\ra}{\rightarrow}
\newcommand{\la}{\leftarrow}

\def\A{{\cal A}}
\def\B{{\cal B}}
\def\C{{\cal C}}
\def\D{{\cal D}}
\def\E{{\cal E}}
\def\F{{\cal F}}
\def\G{{\cal G}}
\def\H{{\cal H}}
\def\I{{\cal I}}
\def\J{{\cal J}}
\def\K{{\cal K}}
\def\L{{\cal L}}
\def\M{{\cal M}}
\def\N{{\cal N}}
\def\O{{\cal O}}
\def\P{{\cal P}}
\def\Q{{\cal Q}}
\def\R{{\cal R}}
\def\T{{\cal T}}
\def\U{{\cal U}}
\def\V{{\cal V}}
\def\W{{\cal W}}
\def\X{{\cal X}}
\def\Y{{\cal Y}}
\def\Z{{\cal Z}}
\def\Re{{\mathbb R}}
\def\Cx{{\mathbb C}}
\def\Ze{{\mathbb Z}}
\def\Na{{\mathbb N}}
\def\ud{\mathrm{d}}
\def\eps{\varepsilon}
\def\dist{\textrm{dist}}



\makeatletter
\renewcommand{\paragraph}{%
  \@startsection{paragraph}{4}%
  {\z@}{1ex}{-1em}%
  {\normalfont\normalsize\bfseries}%
}
\makeatother

%

\title{Audio-driven Neural Gesture Reenactment with Video Motion Graphs
\vspace{-7pt}      
}

\author{ 
        Yang Zhou$^{1,2}$
    \,\,\,\,
        Jimei Yang$^2$
    \,\,\,\,
        Dingzeyu Li$^2$ 
    \,\,\,\,
        Jun Saito$^2$
    \,\,\,\,
        Deepali Aneja$^2$
    \,\,\,\,
        Evangelos Kalogerakis$^1$
    \\
    $^1$University of Massachusetts Amherst \,\,\,\,\,\,\,\,\,\,\,\,\,\,\, $^2$Adobe Research
\vspace{-20pt}    
}

\maketitle

\begin{strip}
\centering
\includegraphics[width=0.95\textwidth]{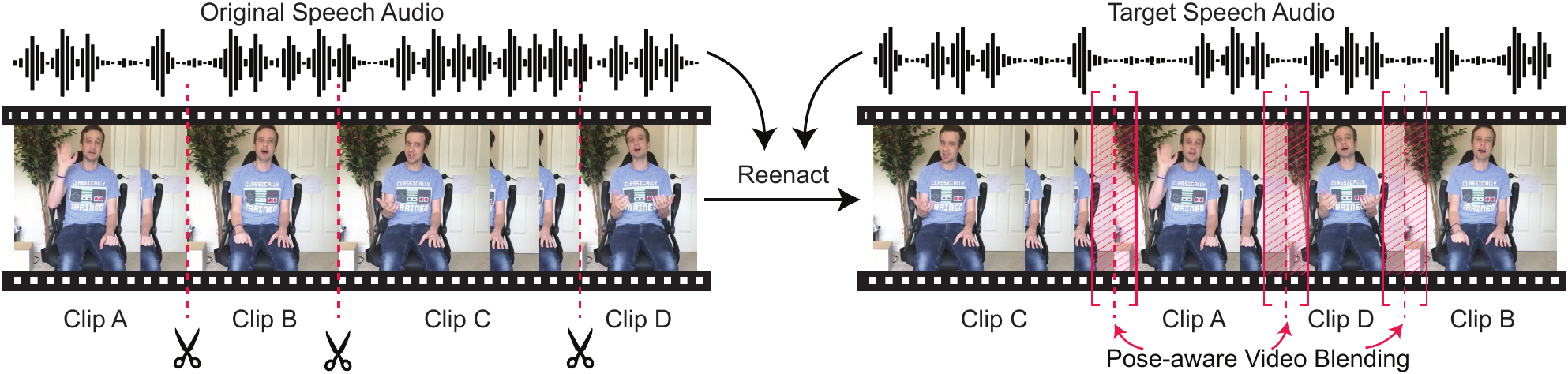}
\vspace{-5pt}
\captionof{figure}
{Given an input reference video of a speaker (left), our method reenacts it with gestures matching a target speech audio (right). The video is synthesized by re-assembling clips from the reference video and blending the inconsistent boundaries with a pose-aware neural network such that the synthesized video is coherent visually and consistent with both the rhythm and content of the target audio.
\label{fig:teaser}}
\end{strip}

\begin{abstract}
\vspace{-7pt}  
Human speech is often accompanied by body gestures including arm and  hand gestures. 
We present a method that reenacts a high-quality video with gestures matching a target speech audio. The key idea of our method is to split and re-assemble clips from a reference video through a novel video motion graph encoding valid transitions between clips. 
To seamlessly connect different clips in the reenactment, we propose a pose-aware video blending network which synthesizes video frames around the stitched frames between two clips. 
Moreover, we developed an audio-based gesture searching algorithm to find the optimal order of the reenacted frames.
Our system generates reenactments that are consistent with both the audio rhythms and the speech content.
We evaluate our synthesized video quality  quantitatively,  qualitatively, and with user studies, demonstrating that our method produces videos of much higher quality and consistency with the target audio compared to previous work and baselines. Our project page \mbox{\url{https://yzhou359.github.io/video_reenact}}
includes code and data.
\vspace{-7pt} 
\end{abstract}


\section{Introduction}

Gesture is a key visual component for human speech communication~\cite{iverson1998people}. 
It enhances the expressiveness of human performance and helps the audience to better comprehend the speech content~\cite{driskell2003effect}.
Given the progress in talking head generation~\cite{zhou2020makelttalk,chen2020talking,thies2020neural,zhou2019talking}, synthesizing plausible gesture videos becomes increasingly important for applications such as digital voice assistants~\cite{naert2020survey} and photo-realistic virtual avatars~\cite{ginosar2019learning,zhou2019talking}.
In this paper, we propose an audio-driven gesture reenactment system that synthesizes speaker-specific human speech video from a target audio clip and a single reference speech video (Figure~\ref{fig:teaser}).

Unlike lip motions with specific phoneme-to-viseme mappings~\cite{edwards2016jali,taylor2017deep,zhou2018visemenet} or facial expressions mostly corresponding to low-frequency sentimental signals~\cite{kaisiyuan2020mead}, gestures exhibit complex relationships with not only acoustics but also semantics of the audio~\cite{mcneill1992hand}. 
Therefore, it is nontrivial to find a direct cross-modal mapping from audio waveform to gesture videos, even for the same speaker.
To bridge the gap between audio and video, previous methods~\cite{ginosar2019learning, liao2020speech2video} predict body pose (i.e., a jointed skeleton) as an intermediate low dimensional representation to drive the video synthesis.
However, they dissect the problem into two independent modules (audio-to-pose, and pose-to-video) and produce results suffering from noticeable artifacts, e.g. distorted body parts and blurred appearance. 

Our method introduces a video reenactment method that is able to synthesize high-resolution, high-quality speech gesture videos directly in the video domain by cutting, re-assembling, and blending clips from a single input reference video. The process is driven by a novel \textit{video motion graph}, inspired by 3D motion graphs used in character animation~\cite{kovar2008motion,arikan2002interactive}.
The graph nodes represent frames in the reference video, and edges encode possible transitions between them. We discover possible valid transitions between frames, and also discover paths in the graph leading to the generation of a new video such that the re-enacted gestures are coherent and consistent with both the audio rhythms and speech content of the target audio. 
 
Direct playback on the discovered paths for an output video can cause temporal inconsistency at the boundary of two disjoint raw frames.
Existing frame blending methods cannot easily solve this problem, especially with fast moving and highly deformed human poses.
Therefore, we also propose a novel human \textit{pose-aware video blending} network to smoothly blend frames around the temporally inconsistent boundaries to produce naturally-looking video transitions. 
By doing so, we successfully transform the problem of audio-driven gesture reenactment into the search for valid paths that best match the given audio.
 
Our path discovery algorithm is motivated by psychological studies on co-speech gesture analysis. The studies show co-speech gestures can be categorized into rhythmic gestures and referential gestures~\cite{mcneill1992hand}.
While rhythmic gestures are well synchronized with audio onsets~\cite{bozkurt2016multimodal,yunus2020sequence}, referential gestures mostly co-occur with certain phrases, e.g. a greeting gesture of hand-waving appears when a speaker says `hello` or `hi`~\cite{cassell2000coordination,bergmann2009gnetic}.
We analyze the speech of the reference video and detect the audio onset peaks~\cite{davis2018visual} as well as a set of keywords from its transcript~\cite{xiong2018microsoft} as audio features added to the corresponding nodes on the video motion graph.
Given the extracted audio onset peaks and keywords from a new audio clip, the optimal paths that best match audio features are used to drive our video synthesis.

Our contributions are summarized as follows:
\vspace{-0.1em}
\begin{itemize}
 \vspace{-0.1em}
    \item a new system that creates high-quality human speech videos with realistic gestures driven by audio only,
\vspace{-0.1em}
    \item a novel video motion graph that preserves the video realism and gesture subtleties,
\vspace{-0.1em}
    \item a pose-aware video blending neural network that synthesizes smooth transitions of two disjoint reference video clips along graph paths, and 
\vspace{-0.1em}
    \item an audio-based search algorithm that drives the video synthesis to match the synthesized gesture frames with both the audio rhythms and the speech content.  
\end{itemize}

\section{Related Work}

\begin{figure*}
    \centering
    \includegraphics[width=0.90\textwidth]{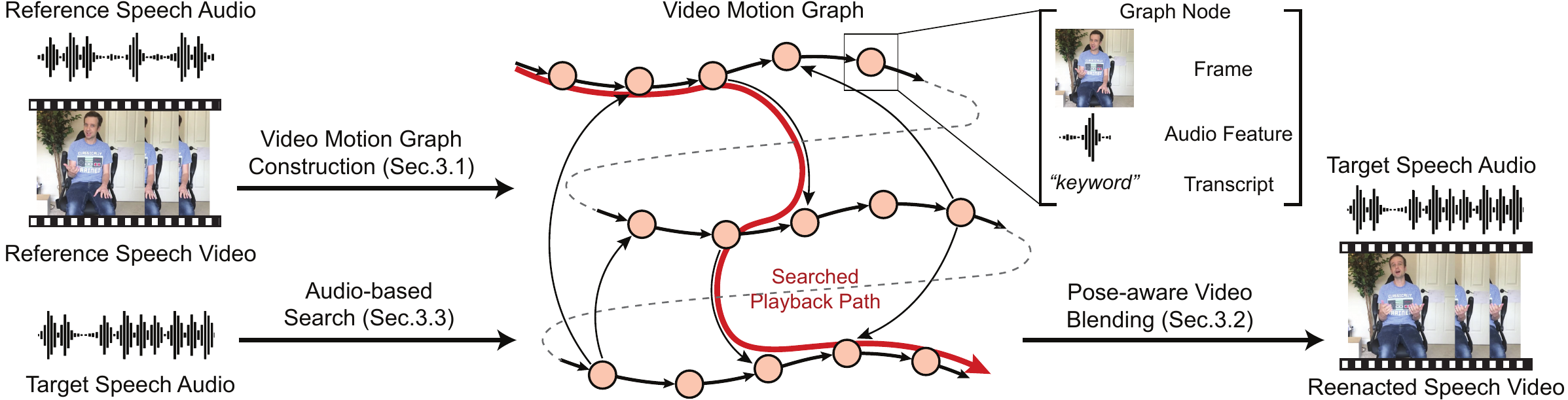}
    \caption{System overview. The reference video is first encoded into a directed graph where nodes represent video frames and audio features, and edges represent transitions. The transitions include original ones between consecutive reference frames, and synthetic ones between disjoint frames.
    Given a unseen target audio at test time, a beam search algorithm finds plausible playback paths such that gestures best match the target speech audio. 
    Synthetic transitions along disjoint frames are neurally blended to achieve temporal consistency.
    \vspace{-7pt}
    }
    \label{fig:overview}
\end{figure*}

Our method is related to previous work on motion graphs, audio-driven 3D speech animation, and in particular human video synthesis, and video frame blending. 

\paragraph{Motion Graph.}
The idea of motion graphs was first proposed in \cite{kovar2008motion,arikan2002interactive} to create realistic and controllable animation based on a pre-captured motion. It is broadly used in generating 3D character animations \cite{heck2007parametric,shin2006fat,beaudoin2008motion,lee2010motion,reitsma2007evaluating,min2012motion,safonova2007construction,kruger2010fast}. However, these approaches only work on 3D human skeleton representations and cannot be directly applied to video animation in image space. While blending re-assembled motions requires interpolating 3D joint positions in character animation, in our case blending requires synthesizing whole image frames to create a coherent video. 

\cite{schodl2000video,agarwala2005panoramic} propose motion graph in pixel space and solve this issue by de-ghosting \cite{shum2001construction} and gradient-domain compositing \cite{wang2004seamless} based on pixel warping. However, these approaches focus on simple periodical scene scenarios, e.g. pendulum, waterfalls, etc. and cannot work on complex human motions. 
\cite{flagg2009human,xu2011video,li2016spa,zhang2020vid2player} generate controllable human action videos by retrieving and warping nearest candidate frames. However, they require additional motion capture resources such as physical markers, multi-view or RGB-D cameras. \cite{casas20144d,casas20154d,huang2015hybrid} also introduce human video synthesis based on reconstruction of human meshes from pre-captured multi-view camera datasets. However, these methods are not suitable for monocular camera videos.

\paragraph{Audio-driven Speech Animation of 3D models.}
Several approaches for 
audio-driven speech animation of lips, 
heads, and body gestures 
have been proposed in the recent years 
\cite{thies2020neural,zhou2020makelttalk,chen2020talking,zhou2019talking,ginosar2019learning, liao2020speech2video}.
\cite{ahuja2020style,kucherenko2020gesticulator,yoon2020speech,alexanderson2020style} propose learning methods to solve the multimodal mapping from audio to 3D human gestures. They represent synthesized gestures with 3D skeletons, which can drive a 3D character model. Yet, these methods are not able to synthesize video of a target speaker unless they are also provided with
a detailed, textured, and rigged 3D model for that speaker. When it is not available, their demonstrated results lack photorealism.

\paragraph{Human Video Synthesis.}
\cite{ginosar2019learning, liao2020speech2video} translate predicted skeletal gesture motions to photo-realistic speaker videos via recent neural image translation approaches \cite{isola2017image,wang2018high,wang2018video,Karras_2019_CVPR}. 
However, neural image translation is not artifact-free: disconnected  moving object parts, as well as incoherent texture appearance are known issues in video generation \cite{wang2018video}.
Due to the large number of parameters in their network, these methods also require large datasets for training.
Few-shot solutions \cite{zakharov2019few,wang2019few} do not have such dataset requirements, yet they suffer from various artifacts, in particular for human pose synthesis, such as blurred appearance and distorted body parts \cite{wang2019few}.
\cite{weng2019photo,liu2019liquid,liu2019neural,shysheya2019textured,weng2020vid2actor}
fit human body model or/and texture parameters to a training video to 
improve the appearance of body shapes and texture at test time. Yet, inaccurate fitting easily results in artifacts and loss of subtleties, especially in the presence of loose clothing and detailed body parts, e.g. fingers. 
\rev{\cite{zheng2019unsupervised,siarohin2018deformable,siarohin2021motion} warp learned features of each body to generate target pose frames based on estimated optical flow. They can handle large pose changes and texture hallucination, but often fail in blending frames naturally.}
Our method follows a largely different approach from all the above prior work: instead of per-frame neural translation, the video of a speaker is generated by re-assembling clips from a short, few minute long reference video. Since most of the frames originate from the reference video,  the synthesized video preserves gesture realism as well as appearance subtleties. As a result, the problem is simplified to blending video frames. Our neural blending network focuses on solving this particular task, instead of generating all frames from scratch.

\paragraph{Video Frame Blending.}
The choice of the frame blending strategy significantly impacts the quality of the video generated from re-assembling clips.  Naive weighted averaging of video frames easily result in ghost effects \cite{schodl2000video,niklaus2020softmax}. More advanced frame interpolation methods \cite{liu2017video,jiang2018super,niklaus2017video,gui2020featureflow} based on optical flow estimation \cite{ilg2017flownet,baker2011database,teed2020raft} have been proposed to synthesize intermediate frames between two consecutive frames, in particular for  slow motion videos. 
However, such methods fail if two frames are very different from each other and the optical flow estimation is not accurate enough. They work for generic content, yet do not consider human motion as a prior for our task. Our method uses a human pose-aware neural network for frame blending that produces significantly better quality video compared to prior such work, as demonstrated in our experiments.

\section{Method}

\label{sec:method}

\paragraph{Overview.}
\label{sec:method_overview}
The goal of our method is to synthesize a new video for a reference speaker given a target speech audio from the same or different speaker. Our video synthesis is guided by a novel \textit{video motion graph} created from an input reference video of the speaker (Sec.~\ref{sec:video_motion_graph}). 
The video motion graph is a directed graph that encodes how the reference video may be split and re-assembled in different graph paths (see Fig.~\ref{fig:overview} for an illustration).
The graph node representations are defined as the raw reference video frames and corresponding audio features. 
The edges are defined as the transitions between frames, including \textit{natural transitions} 
in the input video and \textit{synthetic transitions} connecting disjoint clips. 
Synthetic transitions are introduced to expand the graph connectivity and enable nonlinear video playback. 

However, a direct nonlinear playback along synthetic transitions does not guarantee smooth video rendering due to the abrupt changes of disjoint frames in image space. 
Thus, we design a novel \textit{pose-aware video blending} network  to re-render and interpolate neighboring frames required by the synthetic transitions (Sec.~\ref{sec:blending}).
We develop an \textit{audio-based searching} method to find optimal paths in the video motion graph that best match the target audio features both rhythmically and semantically (Sec.~\ref{sec:searching}). 
To generate new videos, we retrieve the raw input video frames at natural transitions and synthesize neural blended frames at synthetic transitions.

\subsection{Video Motion Graph}
\label{sec:video_motion_graph}

\noindent
The key idea of our video motion graph is to create synthetic transitions based on the similarity of the speaker's pose in the reference video frames.
Our pose similarity metric relies on 3D space and image space cues.
Given a reference video, our first step is to extract pose parameters $\btheta$ of the SMPL model~\cite{loper2015smpl} for all frames with an off-the-shelf motion capture method~\cite{xiang2019monocular}. We further smooth the pose parameters with~\cite{casiez20121} to promote temporally coherent results.

\paragraph{3D space pose similarity.} 
Based on the pose parameters, we compute the 3D positions in world space for all joints via forward kinematics.
For each pair of frames $\forall (m,n)$, we evaluate pose dissimilarity  $d_{feat}(m,n)$  based on the Euclidean distance of their
position and velocity of all joints.

\paragraph{Image space pose similarity.} 
To obtain the pose similarity in image space, for each frame $m$, we project the fitted 3D SMPL human mesh onto image space using known camera parameters from \cite{xiang2019monocular}, 
and mark the mesh surface area which is visible on image after projection as $S_m$.
Then for each pair of frames $(m,n)$, the image space dissimilarity is estimated by 
the Intersection-over-Union (IoU) between their common visible surface areas:~\mbox{$d_{img}(m,n) = 1 - (S_{m} \cap S_{n} ) / (S_{m} \cup S_{n})$}. The lower $d_{img}(m,n)$ is, the higher the IoU, thus larger overlap exists in the surface area in two meshes, indicating higher pose similarity in terms of image rendering.

Based on these two distance measurements, we create graph synthetic transitions between any pair of 
reference video frames (nodes in our graph)
if their distance $d_{feat}(m,n)$ and $d_{img}(m,n)$ are below predefined thresholds (both distance for natural transitions are defined as 0). Here we follow \cite{yang2020statistics} to set the thresholds as the average distance between close frames $(m, m + l)$ in the reference video. Larger frame offset $l$ results in higher thresholds, thus more synthetic transitions, increasing the possible number of paths in the motion graph. This also results in larger computational cost for the path search algorithm of Sec.~\ref{sec:searching}.
Our experiments use $l=4$ which practically achieves a balance between computational cost and number of available paths in the graph.

\subsection{Pose-aware Video Blending}
\label{sec:blending}

A mere playback of connected frames at synthetic transitions easily results in noticeable jittering artifacts (see direct playback in Fig.~\ref{fig:seq_blend}(a) grey dashed path and Fig.~\ref{fig:seq_blend}(b) third column).
To solve this problem, 
we synthesize blended frames to replace original frames around a small temporal neighborhood of a synthetic transition so that the video can smoothly transit from the first sequence to the other (see Fig.~\ref{fig:seq_blend}(a) solid black path and Fig.~\ref{fig:seq_blend}(b) last column). For a synthetic transition connecting frames $m,n$, we define the neighborhood using the frame range $[m-k,m]$ and $[n, n+k]$ with a neighborhood size $k$.

We designed a \textit{pose-aware video blending network} to synthesize frames within the above neighborhood. Given two frames with indices $i,j$ (where $i\in[m-k,m]$ and $j\in[n,n+k]$) and their corresponding raw RGB image representations $I_i$ and $I_j \in \mathbb{R}^{H \times W \times 3}$ from the reference video, the network synthesizes each blended frame in the neighborhood with a target blended weight 
\mbox{
$\alpha \in [0, 1/K, 2/K, ..., 1]$}, where $K=2k$.

As a first step, we use the blending weight to estimate the SMPL pose parameter $\btheta_{t}$ for a blended frame $t$ as:
$\btheta_{t} = (1-\alpha)\btheta_{i} + \alpha\btheta_{j}$,
where $\btheta_{i}$ and $\btheta_{j}$ are the SMPL pose parameters captured from two input frames respectively.

\begin{figure*}[tb]
    \centering
    \includegraphics[width=0.9\textwidth]{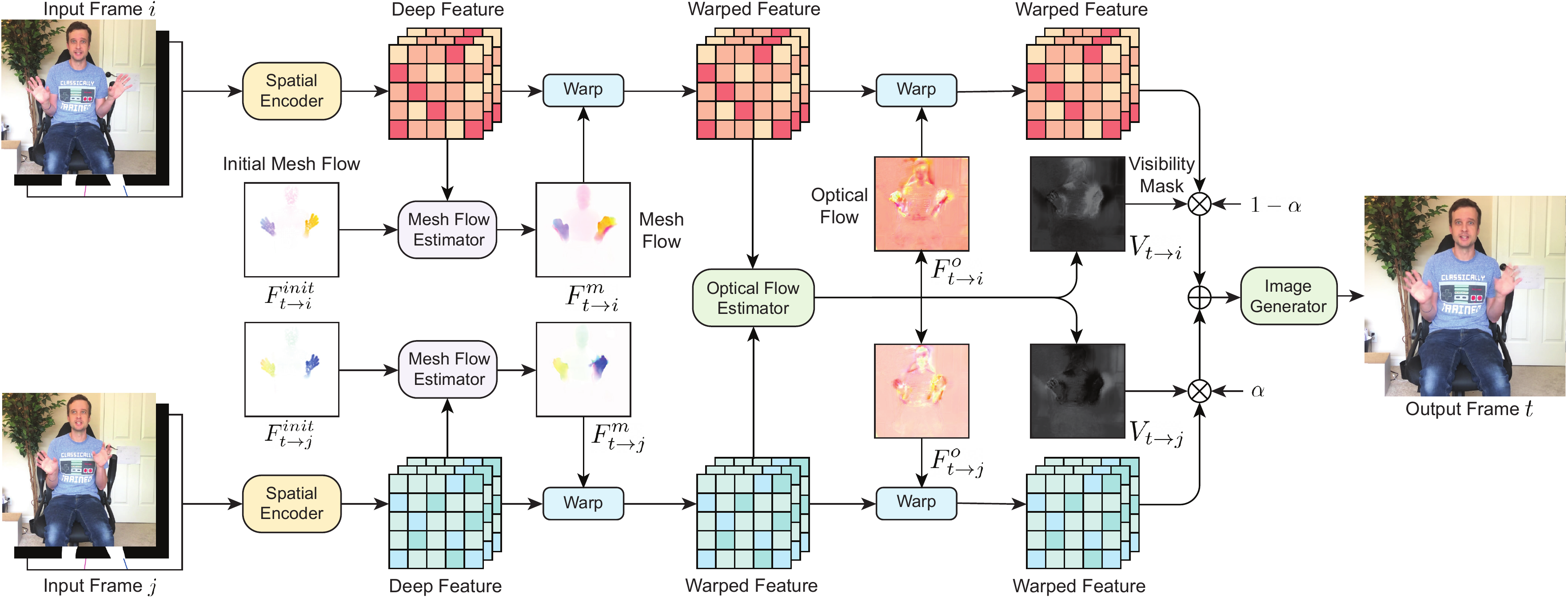}
    \vspace{-2pt}
    \caption{Pose-aware neural blending network architecture. Two source frames are encoded into deep feature maps and then warped based on the predicted flows from two stages: a 3D mesh-based flow stage for coarse feature map alignment, followed by an optical flow-based stage further refining the warping. Finally, the warped features are blended with predicted visibility masks to generate the target frame.}
    \label{fig:frameinterpolation_arch}
    \vspace{-7pt}
\end{figure*}

Our network processes the images   $I_i$,  $I_j$, the body foreground masks,
and the pose parameter $\btheta_{i}, \btheta_{j}, \btheta_{t}$. Processing takes place in two stages. The first stage warps foreground human body image features based on a 3D motion field computed from vertex displacements of the fitted SMPL meshes. The second stage further refines the warping by computing the residual optical flow between the warped image features produced by the first stage, and the optical flow from the rest of the image (i.e., background). Finally, an image translation network transforms the refined warped image features to the image $I_t$ representing
the target output frame $t$. 
The network architecture is shown in Fig.~\ref{fig:frameinterpolation_arch}. 
 
\begingroup
\setlength{\tabcolsep}{1pt} 
\begin{table}[t]
\begin{center}
\begin{tabular}{c}
\includegraphics[width=0.95\linewidth]{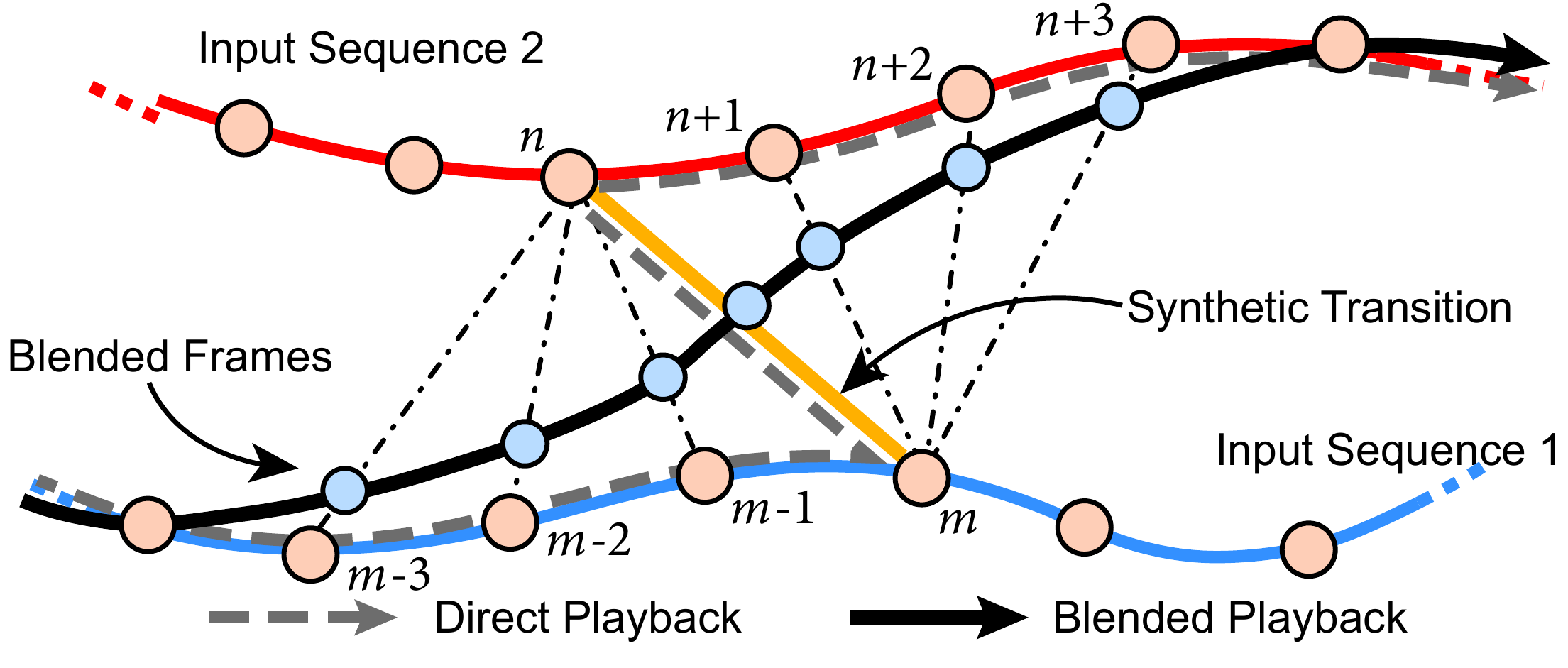} \\
(a) Illustration of pose-aware blended playback. \\
\includegraphics[width=\linewidth]{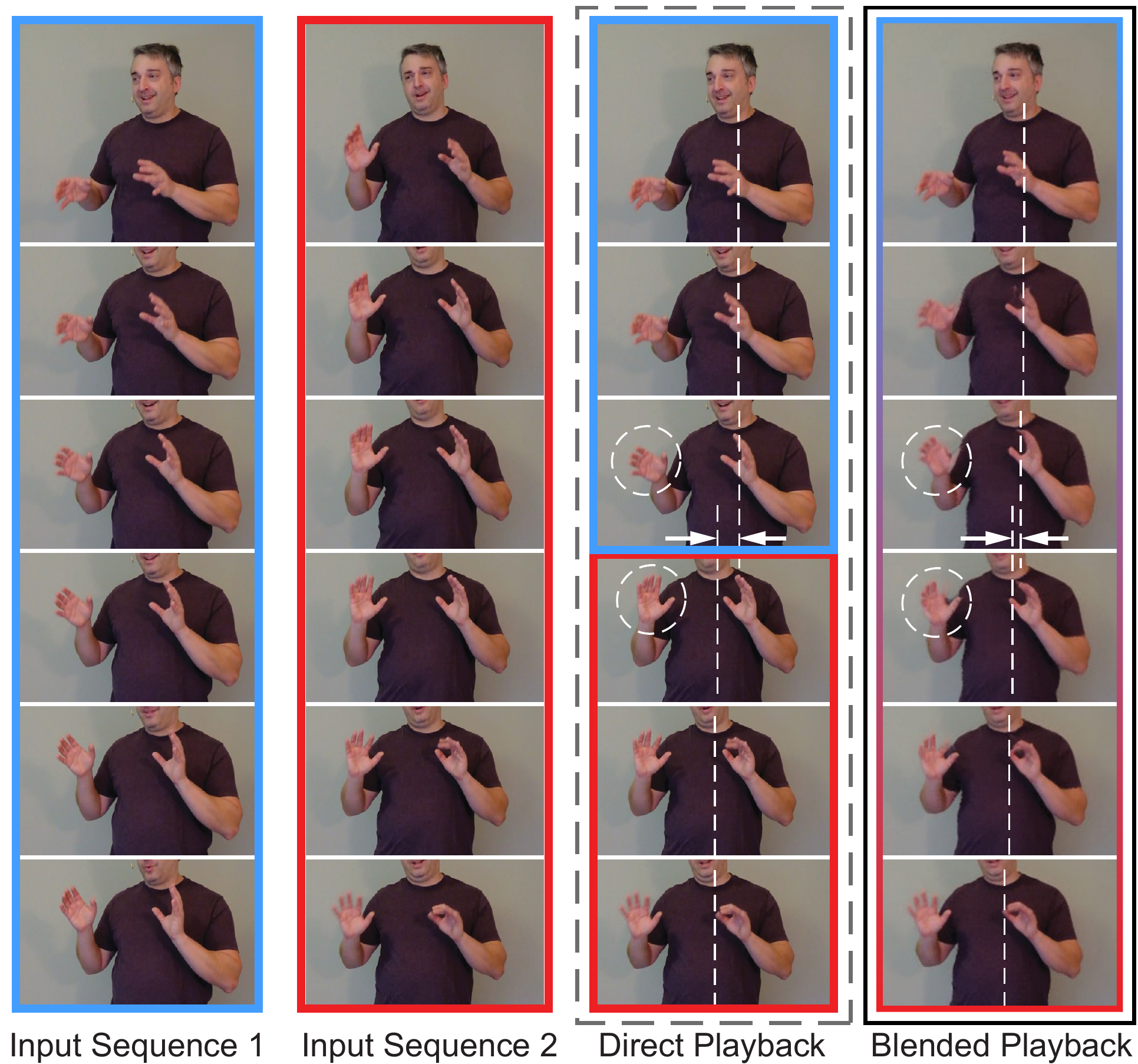} \\ 
(b) Our blended playback generates smoother transition. 
 \\[-1em]
\end{tabular}
\end{center}
\vspace{-2pt}
\captionof{figure}{Compared to direct playback along synthetic transitions which have severe horizontal shift for body poses and abrupt change in hand rotations (see dashed lines and circles in (b)), our blending strategy generates natural transitions between clips. }
\label{fig:seq_blend}
\vspace{-7pt}
\end{table}
\endgroup

\paragraph{Mesh Flow Stage.} The first stage has two parallel streams, each producing image deep feature maps encoding the warping for the input images $I_i$ and $I_j$. To produce these features, we first compute an initial 3D motion field, which we refer to as initial ``mesh flow'', from the SMPL body mesh displacements between the two frames. To this end, we first find the body mesh vertex positions $\bv_i, \bv_j, \bv_t$ from the SMPL pose parameters $\btheta_{i}, \btheta_{j}, \btheta_{t}$ respectively. Then we obtain  the initial mesh flow $F^{init}_{t \to i}$ and $F^{init}_{t \to j}$ as the displacement of the corresponding mesh vertices $\bv_t - \bv_i$ and $\bv_t - \bv_j \in \mathbb{R}^{N \times 3}$ respectively. 
We note that we only consider here the displacements from visible vertices found via perspective projection onto image plane. These  displacements are projected and rasterized as image-space motion field  $\mathbb{R}^{N \times 3} \to \mathbb{R}^{H \times W \times 2}$. Since the vertex sampling does not match the image resolution, the resulting flow fields are rather sparse. Thus, we diffuse them with a Gaussian kernel with $\sigma$ set to $8$ in our experiments. 

These initial motion fields are far from perfect. 
This is because the boundaries of the projected mesh often do not exactly align with the  boundaries of the human body in the input frames.
Thus, we refine these fields with a neural module. 
The module has two streams, each refining the corresponding motion field for frame $i$ and $j$. 
The first stream processes as inputs the RGB image $I_i$, the foreground mask $I_{mask}$, and an image containing the rendered skeleton $I_{skel}$ representing the SMPL pose parameters.
It then encodes them into an image deep feature map $\bx_i$: \begin{equation}
    \bx_i = E_s(I_i, I_{mask}, I_{skel}; \bw_s)
\end{equation}
where $\bw_s$ are  learnable weights. 
Similarly, the second stream produces  an image deep feature map $\bx_j$ for frame $j$. The two streams share the same network based on $8$ stacked CNN residual blocks \cite{brock2018large}. More details are provided in the supplementary material.

We then estimate the refined motion fields through another network $E_m$,
\begin{align}
    F^m_{t\to i} &= E_m(\bx_i, F^{init}_{t \to i}; \bw_m),\\
    F^m_{t\to j} &= E_m(\bx_j, F^{init}_{t \to j}; \bw_m).
\end{align}
where $\bw_m$ are learnable weights.
This network is designed based on UNet~\cite{ronneberger2015u}.
More details are provided in the supplementary material. 
We then backwards warp the above image feature maps with the above motion fields to obtain the warped deep features $\bx'_i$ and $\bx'_j$. 

\begin{figure}
    \centering
    \includegraphics[width=0.45\textwidth]{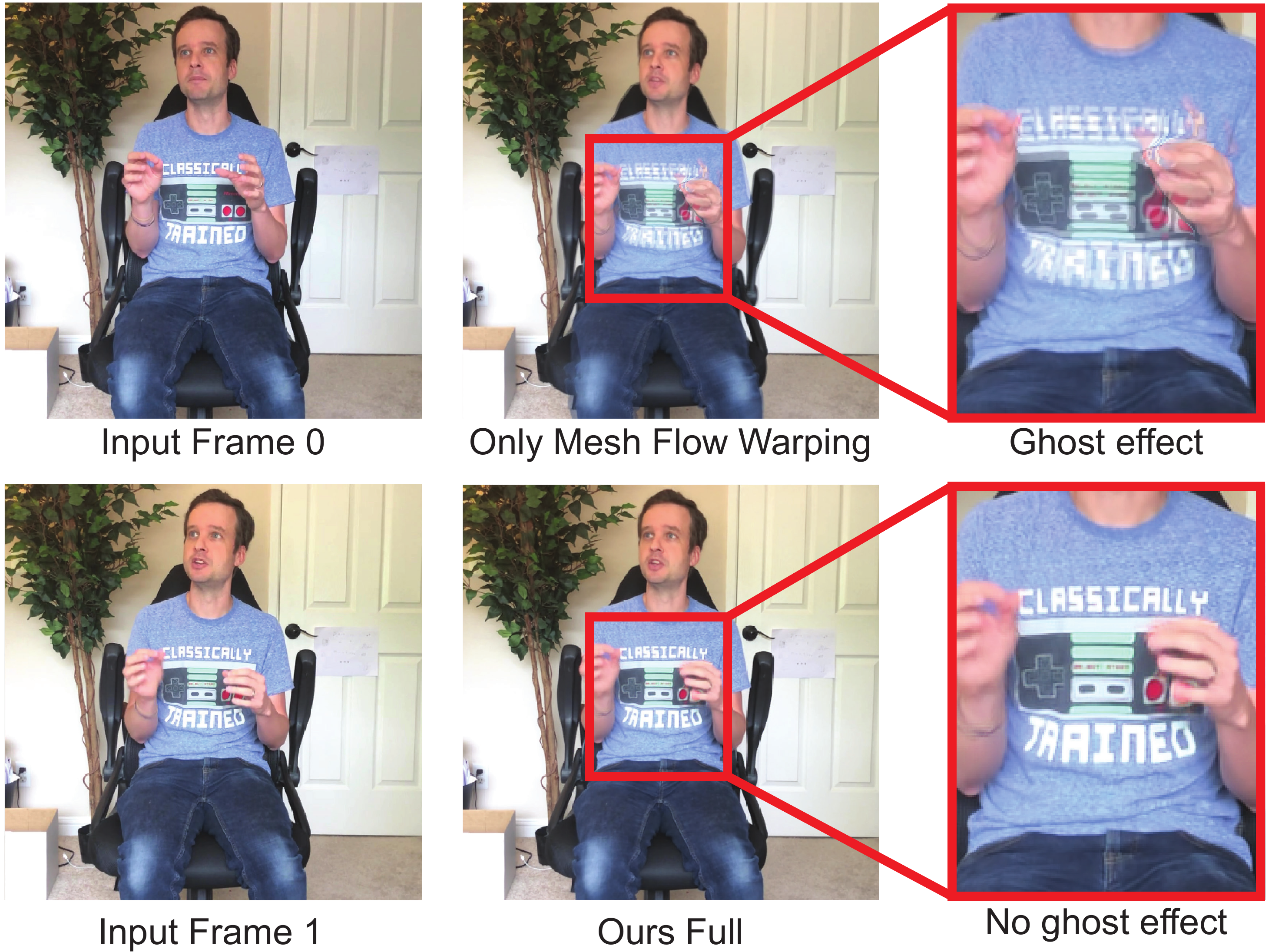}
    \vspace{-2pt}
    \caption{A ghost effect example. \textbf{Left}: two input frames. \textbf{Top-right}: ghost effect from using mesh flow only. \textbf{Bottom-right}: sharp features with further warping by optical flow.}
    \label{fig:ghost_effect}
    \vspace{-7pt}
\end{figure}

\paragraph{Optical Flow Stage.} 
Synthesizing the final target frame directly from the two warped feature maps $\bx'_i$ and $\bx'_j$ suffers from ghost effect (Fig.~\ref{fig:ghost_effect}). This is because the motion field calculated in the previous stage is based on the SMPL model which ignores details such as textures on clothing.

Our second stage aims to further warp the deep feature maps $x'_i$ and $x'_j$ based on optical flow computed throughout the image including the background.
At this stage, the warped features already represent bodies that are roughly aligned.
We found that an off-the-shelf frame interpolation network based on optical flow \cite{jiang2018super} can reproduce the  missing pixel-level details and remedy the ghost effect.
The network predicts optical flow $F^o_{t \to i}$ and $F^o_{t \to j}$ to further warp the features from $\bx'_i$ and $\bx'_j$ to $\bx''_i$ and $\bx''_j$ respectively. It also estimates soft visibility maps \cite{jiang2018super}  $V_{t\to i}$ and $V_{t\to j}$ used for blending to obtain a deep feature map for frame $t$:
\begin{equation}
    x''_t = (1-\alpha)V_{t\to i} \odot x''_i +  \alpha V_{t\to j} \odot x''_j.
\end{equation}
Finally, we take as input the above blended deep feature map to synthesize the target image $I_t$. This is performed with a generator network $G$ following a UNet image translation network architecture \cite{zhou2020makelttalk}: 
%
$\hat{I}_t = G(x''_t; \bw_g)$,
where $\bw_g$ are learnable weights. More details and output examples are provided in the supplementary material. 

\paragraph{Training.} To train  our pose-aware video blending network, we sample triplets of frames in the reference video. Given a target frame e.g., frame $t$, we randomly sample two other nearby frames with indices $t-k_0$ and $t+k_1$, $k_0, k_1 \in [1, 8]$ to form triplets. The corresponding blending weight $\alpha$ is computed as ${k_0}/{(k_0 + k_1)}$. We train the entire network end-to-end with losses defined to better estimate the flows and reconstruct the final image. More details are provided in the supplementary material.

\subsection{Audio-based Search}
\label{sec:searching}

Given a speech audio at test time, we develop a graph search algorithm to find plausible paths along which gestures match the speech audio both rhythmically and semantically. 
Previous studies have shown that speech gestures can be classified into two categories: 1) referential gestures that appear together with specific, meaningful keywords, and 2) rhythmic gestures which respond to the audio prosody features~\cite{mcneill1992hand}. 
More specifically, the key stroke of a rhythmic gesture appear at the same time as (or within a very short of period of) an \textit{audio onset} within a phonemic clause \cite{elhoseiny2017sherlock}. 
To find precise gestures on the right timings, or frame indices, we define  a pair of audio features for input speech: \emph{audio onset feature} and \emph{keyword feature}. 
The audio frame indices match the video frame rate.

\paragraph{Audio onset and keyword feature.} We define the audio onset feature as a binary value indicating the activation of an audio onset for each frame detected with a standard audio processing algorithm~\cite{bello2005tutorial}. 
To extract keyword features, 
we first use the Microsoft Azure speech-to-text engine \cite{xiong2018microsoft} to convert the input audio into transcripts with corresponding start and end time 
for each word. 
We create a dictionary of common words for referential gestures, which we call \emph{keywords}  (see supplementary for a list). 
If a keyword appears at a frame (or node), we set its keyword feature to that word. 
Otherwise, we simply set it to \textit{empty} (no keyword).

\begingroup
\setlength{\tabcolsep}{1pt} 
\begin{table*}
\begin{center}
\begin{tabular}{c c c c c c c c}
\includegraphics[width=0.1320\linewidth]{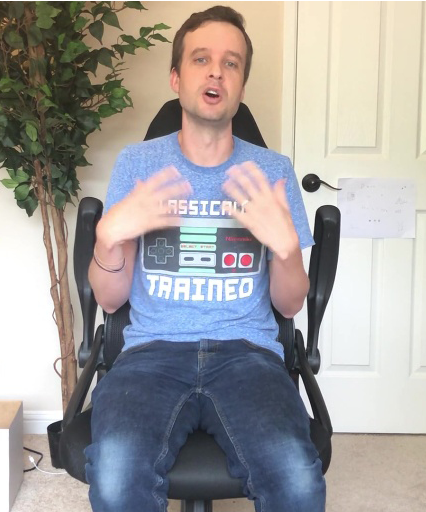} &
\includegraphics[width=0.1320\linewidth]{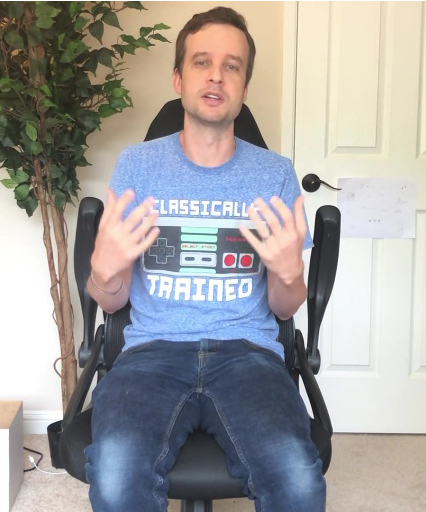} & &
\includegraphics[width=0.1320\linewidth]{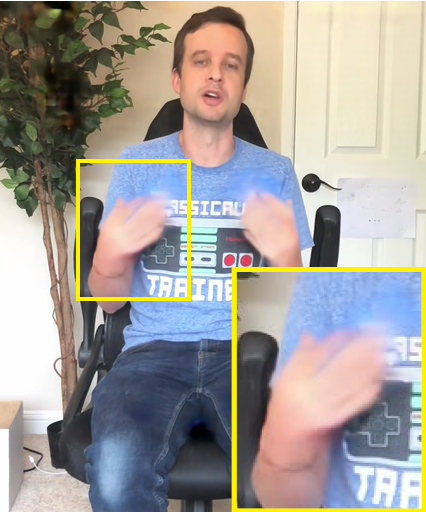} &
\includegraphics[width=0.1320\linewidth]{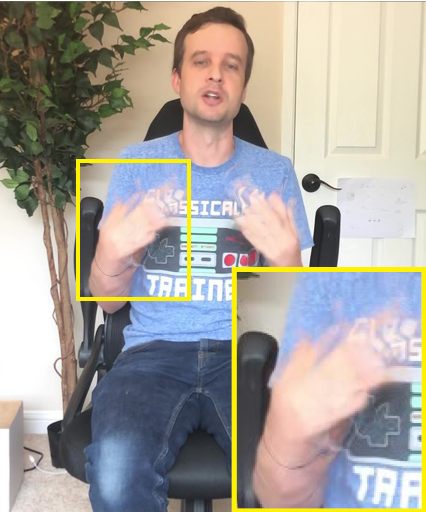} &
\includegraphics[width=0.1320\linewidth]{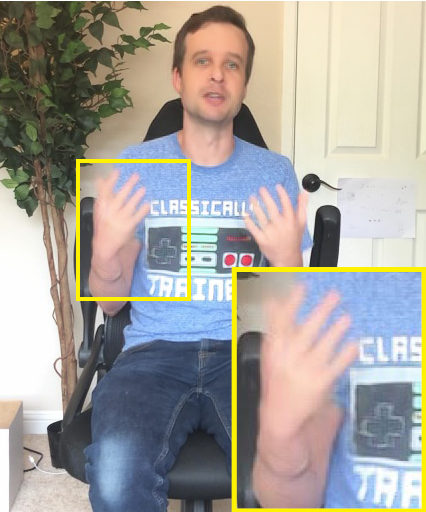} &
\includegraphics[width=0.1320\linewidth]{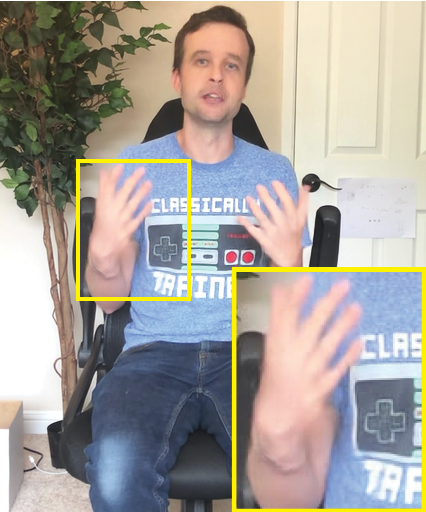} &
\includegraphics[width=0.1320\linewidth]{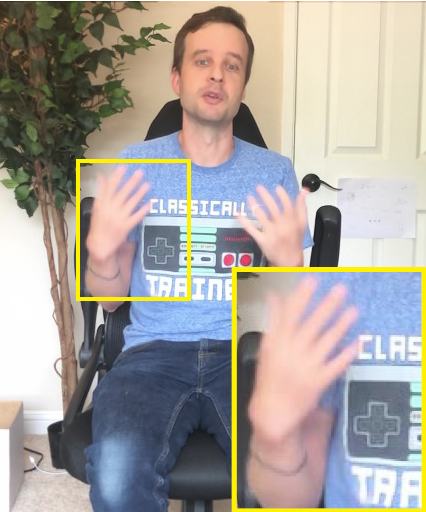} \\


\includegraphics[width=0.1320\linewidth]{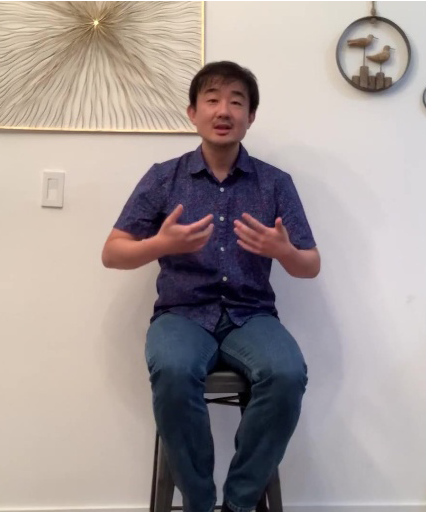} &
\includegraphics[width=0.1320\linewidth]{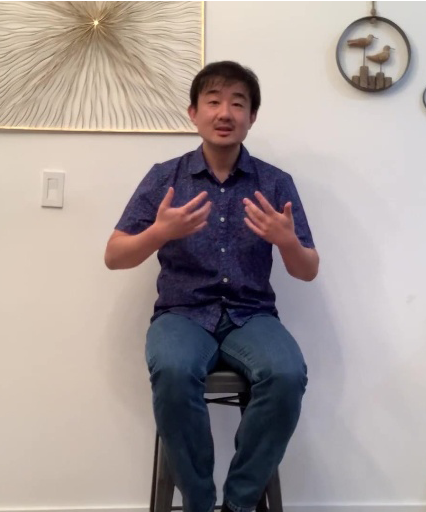} & &
\includegraphics[width=0.1320\linewidth]{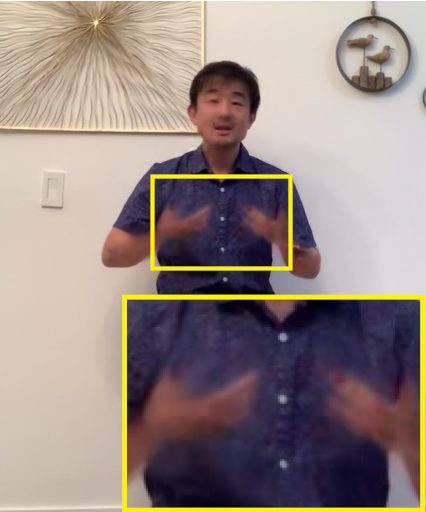} &
\includegraphics[width=0.1320\linewidth]{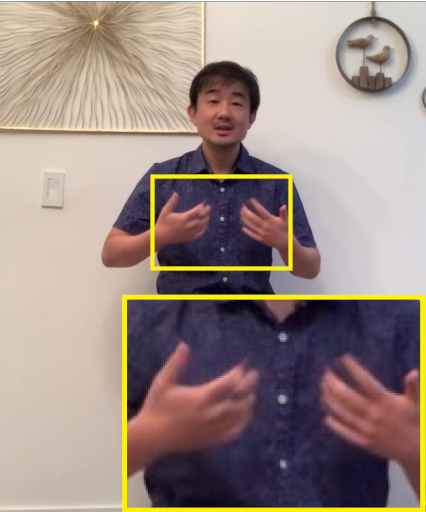} &
\includegraphics[width=0.1320\linewidth]{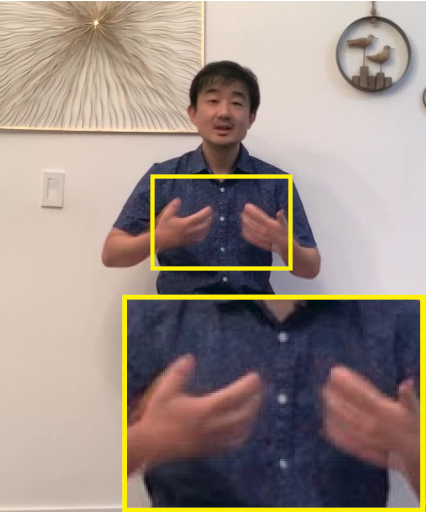} &
\includegraphics[width=0.1320\linewidth]{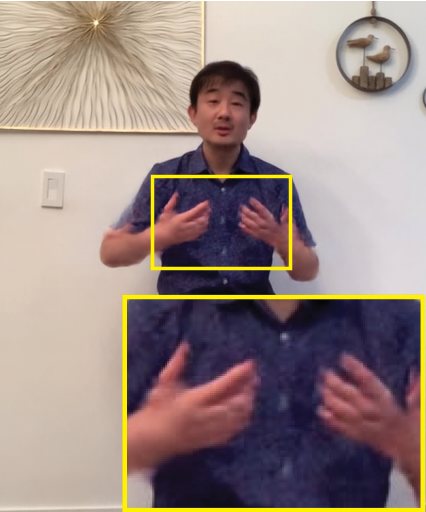} &
\includegraphics[width=0.1320\linewidth]{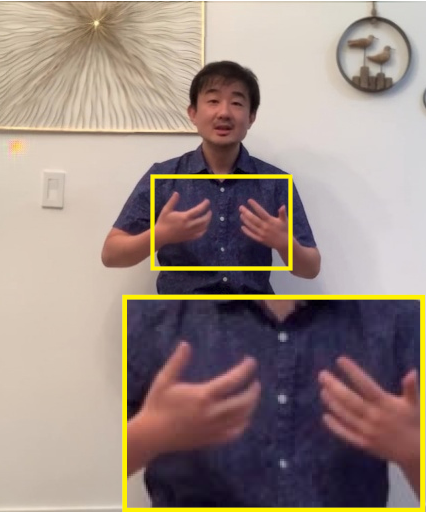} \\

\multicolumn{2}{c}{2 Input Frames} & \ &
 FeatureFlow\cite{gui2020featureflow} & SuperSlMo\cite{jiang2018super} & vUnet\cite{esser2018variational} & EBDance\cite{chan2019dance} & Ours\\
 \\[-2.5em]
\end{tabular}
\end{center}
\vspace{-2pt}
\captionof{figure}{Comparison of blended frame synthesis using different methods. Note the natural look of details such as fingers in our method.}
\label{fig:interpolation_compare}
\vspace{-2pt}
\end{table*}
\endgroup

\begingroup
\setlength{\tabcolsep}{1pt} 
\begin{table*}
\begin{center}
\begin{tabular}{c c c c c c c c}

\includegraphics[width=0.116\linewidth]{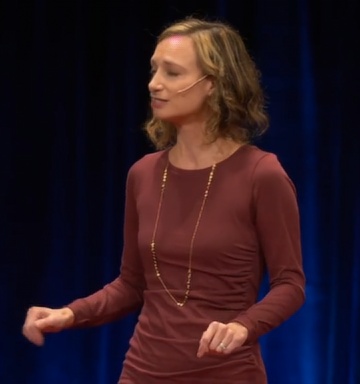} &
\includegraphics[width=0.116\linewidth]{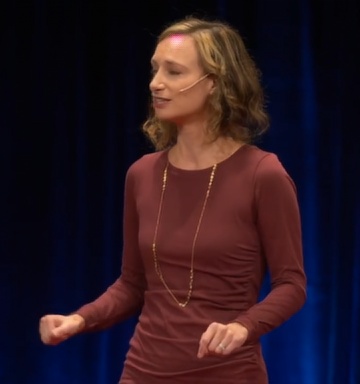} &
\includegraphics[width=0.116\linewidth]{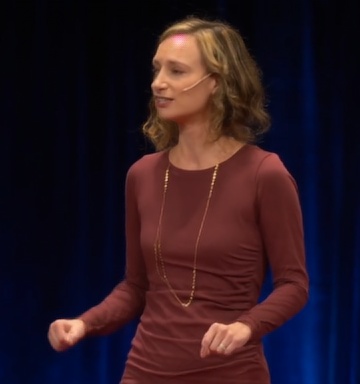} &
\includegraphics[width=0.116\linewidth]{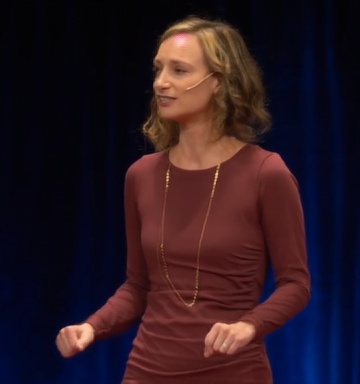} &
\includegraphics[width=0.116\linewidth]{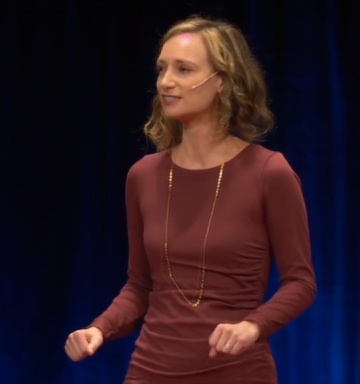} &
\includegraphics[width=0.116\linewidth]{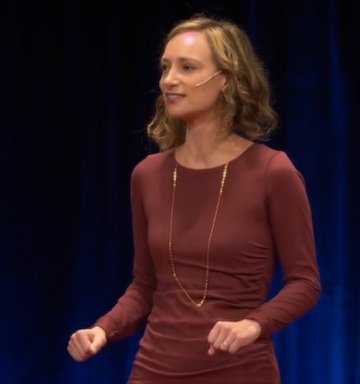} &
\includegraphics[width=0.116\linewidth]{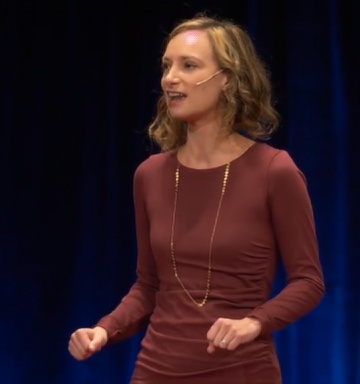} &
\includegraphics[width=0.116\linewidth]{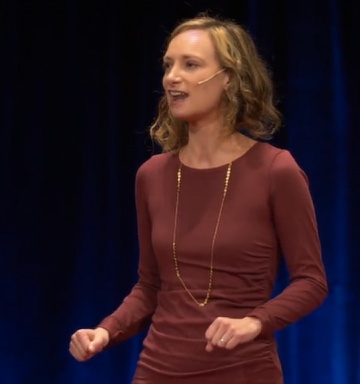} \\


\includegraphics[width=0.116\linewidth]{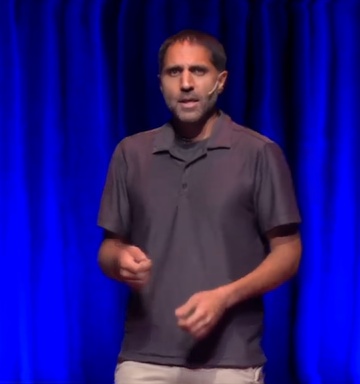} &
\includegraphics[width=0.116\linewidth]{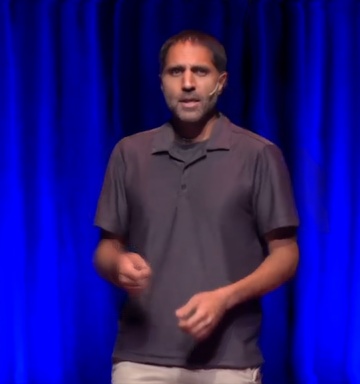} &
\includegraphics[width=0.116\linewidth]{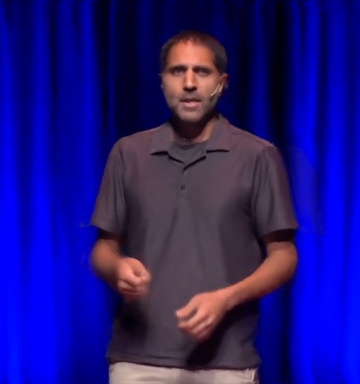} &
\includegraphics[width=0.116\linewidth]{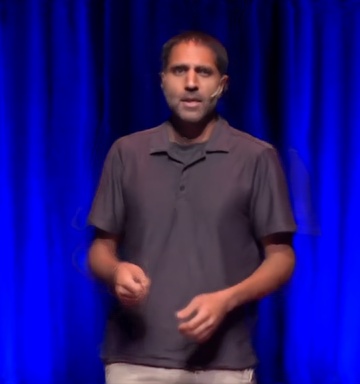} &
\includegraphics[width=0.116\linewidth]{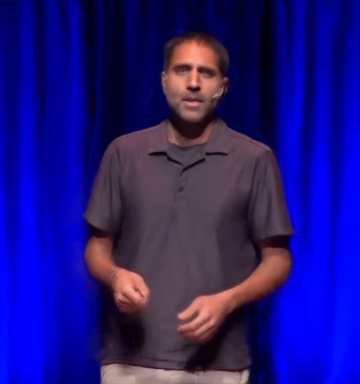} &
\includegraphics[width=0.116\linewidth]{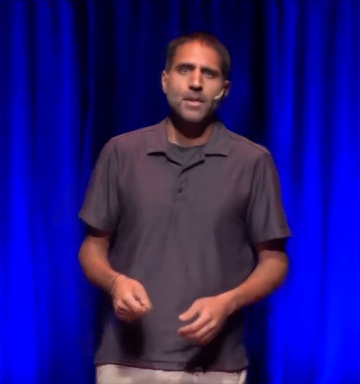} &
\includegraphics[width=0.116\linewidth]{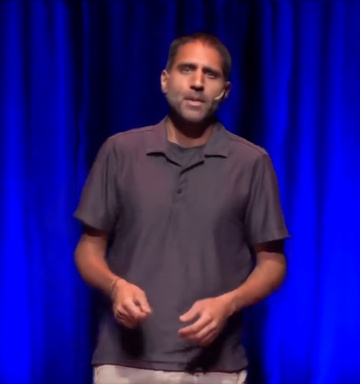} &
\includegraphics[width=0.116\linewidth]{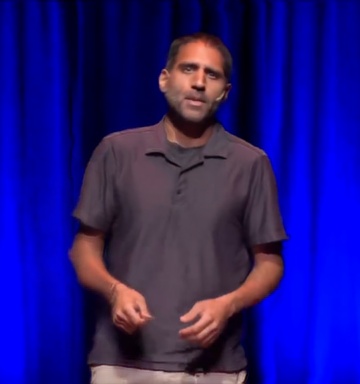} \\

Input $I_i$ & \multicolumn{6}{c}{Synthesized blended frames to smoothly transit from frame $I_i$ to $I_j$} & Input $I_j$\\
 \\[-2.5em]
\end{tabular}
\end{center}
\vspace{-2pt}
\captionof{figure}{Blended frames for transition edges of our video motion graph on TED-talks dataset (see demo videos on our project page).}
\label{fig:ted_result}
\vspace{-7pt}
\end{table*}
\endgroup

\paragraph{Target speech audio segmentation.} 
We split the target speech audio into segments starting and ending with the frames where the audio onset or keyword feature is activated.
Let $\{a_s\}_{s=1}^S$ be the frame indices of such frames, where $S$ is their total number. 
Segments are represented as $a_s \to a_{s+1}$, and their duration are $L_s = a_{s+1}-a_s$ (number of frames).
We also add two extra endpoints $a_0=1$ and $a_{S+1}=N_t$ indicating the first and last frame of the target audio respectively to form the complete segment list, i.e. $a_s \to a_{s+1}$, $s=0,1,..,S$.

\paragraph{Beam search.} 
We utilize the beam search \cite{rubin1977locus} in the video motion graph to find $K$ plausible paths matching the target speech audio segments.
The beam search initializes $K$ paths starting with $K$ random nodes as the first frame $a_0$ for the target audio. \revise{Next, we apply breadth-first search to find path segments ending on nodes whose feature matches the target audio segment feature at frame $a_1$. We continue with the same search procedure as above to find full graph paths matching the rest of the target segments $a_s \to a_{s+1}$, $s=1,..,S$ iteratively. All searched $K$ paths can be used to generate various plausible results for the same target speech audio. Detailed search criteria and result variants can be found in the supplementary material and our project page.
}

\paragraph{Video synthesis.}
\label{sec:synthesis}
 We generate a video along with the final path in the motion graph discovered by the  beam search executions, and use the blending network to handle synthetic transitions (see Fig.~\ref{fig:seq_blend} for an example).  
 As explained above, for each synthesized video segment corresponding to target audio segment $a_s \to a_{s+1}$, we adjust its speed to match the target duration.
Finally, we post-process our result by adopting \cite{prajwal2020lip} to synchronize the lips of the speaker to match the corresponding speech audio.




\section{Results and Evaluation}
\revise{
\paragraph{Dataset.} We evaluate our neural blending network and produced audio-driven reenactment results on two datasets. 

\textit{Personal Story Dataset.} 
Since our approach works for speaker-specific speech gesture reenactment, we collected seven speech videos. 
Each speaker is asked to tell a personal story in front of a static camera, either standing or sitting.
Speakers are encouraged to use their gestures while telling the stories.
The length of the video varies between 2-10 minutes depending on the story.
We split each video into $90\%/10\%$ for training and testing purpose.

\textit{TED-talks dataset \cite{siarohin2021motion}.} We also demonstrate the generalization of our neural blending network on the TED-talks dataset. It contains $1265$ talk speech videos with $393$ unique speakers. Each video contains the upper part of speaker body and the video length ranges from $2$ to $60$ seconds. We use the same train/test split proposed in \cite{siarohin2021motion}. 
\rev{We evaluated the generalization ability of our model on this dataset since the test speakers are unseen during training.}
}

\subsection{Video Blending Evaluation}
\label{sec:vb-evaluation}
We firstly numerically evaluate the proposed video blending network on both dataset. 
Given two frames $t-k$ and $t+k$ in the test split of each video, we synthesize blended frames with the blending weight $\alpha=0.5$ and compare its quality with the ground-truth frame $t$.
All the compared frames are multiplied with ground-truth human masks to compare the foreground human results only.

We compare our method with the state-of-the-art frame interpolation methods \textbf{FeatureFlow}\cite{gui2020featureflow} and \textbf{SuperSlMo}\cite{jiang2018super}, as well as human pose-based image synthesis methods \textbf{vUnet}\cite{esser2018variational}. 
\revise{We also compare with methods based on the 
pix2pix \cite{wang2018high} backbone: the \textbf{EBDance}\cite{chan2019dance} method for speaker-specific Personal story dataset and the \textbf{Fewshot-vid2vid}\cite{wang2019few} for the speaker-varying TED-talks dataset.}
For the pose-based image synthesis methods, we interpolate human skeleton by averaging joint positions.
We retrain all the comparison methods on our dataset for a fair comparison.
We also evaluate two network alternatives: \textbf{Ours w/ mesh} which only uses mesh-based warping flows and \textbf{Ours w/ optical} which only uses optical flows.

\paragraph{Image Quality.}

We evaluate the quality of synthesized images via four common metrics: Image Error (IE) - average absolute pixel difference between two images; Peak Signal-to-Noise Ratio (PSNR) and LPIPS \cite{zhang2018unreasonable}.

\begin{table*}
\begin{center}
\begin{tabular}{c c c c c c | c c c c c}
\hline
 & \multicolumn{5}{c}{Personal story dataset} & \multicolumn{5}{|c}{TED-talks dataset} \\
\hline

Method & IE$\downarrow$ & PSNR$\uparrow$ & LPIPS$\downarrow$ & MOVIE$\downarrow$ & FID$\downarrow$ & IE$\downarrow$ & PSNR$\uparrow$ & LPIPS$\downarrow$ & MOVIE$\downarrow$ & FID$\downarrow$ \\
\hline\hline
FeatureFlow\cite{gui2020featureflow} & 1.18 & 33.5 & 0.015 & 0.22 & 19.1 & 5.2 & 19.7 & 0.267 & 1.29 & 33.6\\
SuperSlMo\cite{jiang2018super} & 1.04 & 35.0 & 0.012 & 0.17 & 15.4 & 1.18 & 28.6 & 0.052 & 0.50 & 12.6\\
vUnet\cite{esser2018variational} & 1.20 & 33.6 & 0.013 & 0.19 & 15.6 & 1.19 & 28.8 & 0.058 & 0.52 & 14.0 \\
EBDance\cite{chan2019dance} & 1.75 & 30.7 & 0.020 & 0.43 & 20.5 & - & - & - & - & -\\
Fewshot-vid2vid\cite{wang2019few} & - & - & - & - & - & 10.7 & 15.1 & 0.159 & 1.06 & 21.5\\
\hline
Ours w/ mesh & 0.87 & 35.2 & 0.009 & 0.14 & 15.1 & 1.36 & 27.9 & 0.072 & 0.64 & \textbf{11.5}\\
Ours w/ optical & 0.97 &34.6 & 0.009 & 0.16 & 13.2 & 1.25 & 28.2 & 0.069 & 0.57 & 11.9\\
Ours & \textbf{0.76} & \textbf{36.1} & \textbf{0.007} & \textbf{0.13} & \textbf{13.0} & \textbf{0.93} & \textbf{30.7} & \textbf{0.040} & \textbf{0.43} & 11.8 \\
\hline\\[-1.8em]
\end{tabular}
\end{center}
\vspace{-7pt}
\caption{Image and video quality assessment for Personal story dataset and TED-talks dataset.}
\label{table:comparison}
\vspace{-7pt}
\end{table*}

\rev{Table \ref{table:comparison} shows our model consistently outperforms all other methods for speaker-specific videos on the Personal story dataset. It also demonstrates the generalization of our model for unseen speakers on the TED-talks dataset.}
Fig.~\ref{fig:interpolation_compare} shows examples of synthesized frames by different methods.
In the top example, the inputs are two frames with larger gesture difference.
The frame interpolation methods \cite{gui2020featureflow,jiang2018super} cannot estimate the flow field, and thus result in broken and blurred hand results.
The pose-based image synthesis methods \cite{esser2018variational,chan2019dance} preserve hand structures but have artifacts around fingers and clothing. 
Ours achieves the best quality for both hands and clothing. 
The lower example shows frames with smaller gesture differences. \cite{jiang2018super,esser2018variational,chan2019dance} preserve hands better but still suffer from broken and blurred texture.
Ours generates clear and sharp results.

\paragraph{Video Quality.} 
To evaluate the quality of the generated video, we adopt the metric, MOVIE \cite{seshadrinathan2009motion} index, to evaluate the video distortion in spatio-temporal aspects.
We also follow \cite{wang2018video} to evaluate the visual quality of the video and temporal consistency with Fréchet Inception Distance (FID) scores~\cite{heusel2017gans}. 
We use the pre-trained video recognition CNN model to get features from synthesized video clips~\cite{carreira2017quo}.
Table~\ref{table:comparison} relative columns 
show our method can achieve the best video quality in the temporal domain.
It demonstrates that the synthesized blended frames  seamlessly connect reenacted frames with much less temporal artifacts.
\revise{In Fig.\ref{fig:ted_result}, we show detailed blended frames on the selected transition edges of our video motion graph from the TED-talks dataset. 
We provide additional synthesized clips to showcase  blending results on our project page.
}

\begin{figure}
    \centering
    \includegraphics[width=0.45\textwidth]{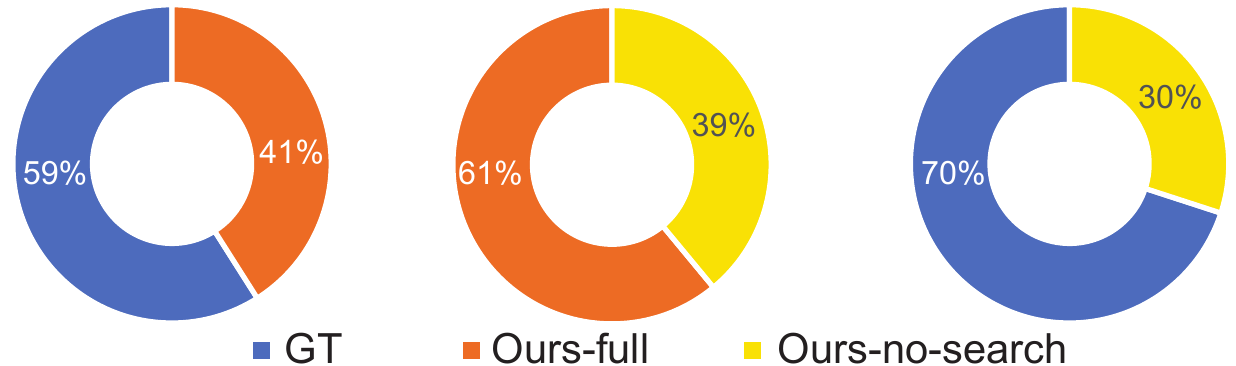}
    \caption{Pairwise comparison results from our user study. The comparison of \textit{Ours-full} against \textit{Ours-no-search} shows the effectiveness of the proposed audio-based search algorithm.
    }
    \vspace{-7pt}
    \label{fig:user_study}
\end{figure}

\subsection{Audio-driven Reenactment Results}
\label{sec:system-evaluation}
\revise{
Given a reference video from speaker $A$ and a target audio clip randomly from another speaker $B$, 
we can reenact the reference video to generate a new speech video with $A$'s appearance and $B$'s voice based on our pipeline. The reenacted results on both the Personal story dataset and the TED-talks dataset are provided on the project page.
}

\paragraph{User Study.} 
\revise{To further quantitatively evaluate the consistency of such reenacted videos to target speech audio, we perform a perceptual user study on the reenacted videos on the Personal story dataset. We generate $127$ such videos of $25$ seconds in length. Each of them contains expressive speech gestures for every speaker in the dataset. The study was conducted via the Amazon Mechanical Turk service.  
} 
We compare the results from our full system (\textbf{Ours-full}) against ground-truth (\textbf{GT}), which are original reference video clips of speaker $B$, and results from a baseline system (\textbf{Ours-no-search}), which randomly finds paths along video motion graphs without audio-based search.

We design the user study questionnaire by providing a list of queries involving pairwise comparisons of results from two out of three methods mentioned above.
\revise{The participants were asked to choose which gestures in those two results are more consistent with the speech audio. Detailed setup to prevent biases and invalid answers can be found in the supplementary material. Finally, 1130 valid choices from 113 valid participants are gathered. We plot the statistics in Fig.~\ref{fig:user_study}. The preference ($61\%$ vs. $39\%$) of \textit{Ours-full} over \textit{Ours-no-search} shows the effectiveness of the audio-based search algorithm.
Although no audio guidance is used, $30\%$ votes received by \textit{Ours-no-search} against \textit{GT} also suggest our video motion graph and frame blending approach is able to generate high-quality and realistic videos. 
The relative higher votes ($41\%$) given to \textit{Ours-full} against \textit{GT} demonstrates our full system generates better though not perfect gesture videos that are coherent with the audio.}




\section{Conclusion and Future Work}

\noindent
We propose a novel system 
based on video motion graphs to generate new videos that best preserve high image synthesis quality and speaker gesture motion subtleties. 
To seamlessly reenact disjoint frames from the input video, we introduce a neural pose-aware video blending method to smoothly blend inconsistent transition frames. 
We show the superior performance of the proposed system comparing to the state-of-the-art methods and baselines via both numerical experiments and perceptual user studies. 

\paragraph{Limitations.}
We use a pre-defined common keyword dictionary 
for keyword features, which may fail on uncommon individual vocabulary.
Using
richer audio features learnt through data might help with accurate gesture matching.
There is an inevitable trade-off between the quality and variety of synthesized animations: increasing the graph edge density can increase the transition variety, yet may retrieve frames harder to blend. 
The proposed video blending network can blend the foreground human poses and slight background changes, but it fails on dramatically changed backgrounds (see the supplementary for examples).

\paragraph{Future work.} Neural blending shows its strengths on reenacting human videos in the pose-aware embedding space.
We believe our hybrid framework of video motion graph and neural reenactment is a promising direction for high-quality controllable digital human animations. 

\paragraph{Potential negative societal impacts.}
Our approach enables synthesis of talking people. This offers the ability of creating fake videos for malicious purposes. Detecting deep fakes videos \cite{rossler2019faceforensics++,li2020celeb,zhou2021face} is an active area of research.

\paragraph{Acknowledgements.} Our research was partially funded by NSF (EAGER-1942069) and Adobe.

{\small
\bibliographystyle{ieee_fullname}
\bibliography{egbib}

\begin{thebibliography}{10}\itemsep=-1pt

\bibitem{agarwala2005panoramic}
Aseem Agarwala, Ke~Colin Zheng, Chris Pal, Maneesh Agrawala, Michael Cohen,
  Brian Curless, David Salesin, and Richard Szeliski.
\newblock Panoramic video textures.
\newblock In {\em ACM Trans. on Graphics (TOG)}. 2005.

\bibitem{ahuja2020style}
Chaitanya Ahuja, Dong~Won Lee, Yukiko~I Nakano, and Louis-Philippe Morency.
\newblock Style transfer for co-speech gesture animation: A multi-speaker
  conditional-mixture approach.
\newblock In {\em Proc. ECCV}, 2020.

\bibitem{alexanderson2020style}
Simon Alexanderson, Gustav~Eje Henter, Taras Kucherenko, and Jonas Beskow.
\newblock Style-controllable speech-driven gesture synthesis using normalising
  flows.
\newblock In {\em Computer Graphics Forum}, 2020.

\bibitem{arikan2002interactive}
Okan Arikan and David~A Forsyth.
\newblock Interactive motion generation from examples.
\newblock {\em ACM Trans. on Graphics (TOG)}, 2002.

\bibitem{baker2011database}
Simon Baker, Daniel Scharstein, JP Lewis, Stefan Roth, Michael~J Black, and
  Richard Szeliski.
\newblock A database and evaluation methodology for optical flow.
\newblock {\em IJCV}, 2011.

\bibitem{beaudoin2008motion}
Philippe Beaudoin, Stelian Coros, Michiel van~de Panne, and Pierre Poulin.
\newblock Motion-motif graphs.
\newblock In {\em Proc. ACM SCA}, 2008.

\bibitem{bello2005tutorial}
Juan~Pablo Bello, Laurent Daudet, Samer Abdallah, Chris Duxbury, Mike Davies,
  and Mark~B Sandler.
\newblock A tutorial on onset detection in music signals.
\newblock {\em IEEE Trans on Speech and Audio Processing}, 2005.

\bibitem{bergmann2009gnetic}
Kirsten Bergmann and Stefan Kopp.
\newblock Gnetic--using bayesian decision networks for iconic gesture
  generation.
\newblock In {\em International Workshop on Intelligent Virtual Agents}, 2009.

\bibitem{bozkurt2016multimodal}
Elif Bozkurt, Y{\"u}cel Yemez, and Engin Erzin.
\newblock Multimodal analysis of speech and arm motion for prosody-driven
  synthesis of beat gestures.
\newblock {\em Speech Communication}, 2016.

\bibitem{brock2018large}
Andrew Brock, Jeff Donahue, and Karen Simonyan.
\newblock Large scale gan training for high fidelity natural image synthesis.
\newblock In {\em Proc. ICLR}, 2018.

\bibitem{carreira2017quo}
Joao Carreira and Andrew Zisserman.
\newblock Quo vadis, action recognition? a new model and the kinetics dataset.
\newblock In {\em Proc. CVPR}, 2017.

\bibitem{casas20154d}
Dan Casas, Christian Richardt, John Collomosse, Christian Theobalt, and Adrian
  Hilton.
\newblock 4d model flow: Precomputed appearance alignment for real-time 4d
  video interpolation.
\newblock In {\em Computer Graphics Forum}, 2015.

\bibitem{casas20144d}
Dan Casas, Marco Volino, John Collomosse, and Adrian Hilton.
\newblock 4d video textures for interactive character appearance.
\newblock In {\em Computer Graphics Forum}, 2014.

\bibitem{casiez20121}
G\'{e}ry Casiez, Nicolas Roussel, and Daniel Vogel.
\newblock 1 € filter: A simple speed-based low-pass filter for noisy input in
  interactive systems.
\newblock In {\em Proc. SIGCHI on Human Factors in Computing Systems}, 2012.

\bibitem{cassell2000coordination}
Justine Cassell, Matthew Stone, and Hao Yan.
\newblock Coordination and context-dependence in the generation of embodied
  conversation.
\newblock In {\em Proc. International Conference on Natural Language
  Generation}, 2000.

\bibitem{chan2019dance}
Caroline Chan, Shiry Ginosar, Tinghui Zhou, and Alexei~A Efros.
\newblock Everybody dance now.
\newblock In {\em Proc. ICCV}, 2019.

\bibitem{chen2020talking}
Lele Chen, Guofeng Cui, Celong Liu, Zhong Li, Ziyi Kou, Yi Xu, and Chenliang
  Xu.
\newblock Talking-head generation with rhythmic head motion.
\newblock In {\em Proc. ECCV}, 2020.

\bibitem{davis2018visual}
Abe Davis and Maneesh Agrawala.
\newblock Visual rhythm and beat.
\newblock {\em ACM Trans. on Graphics (TOG)}, 2018.

\bibitem{driskell2003effect}
James~E Driskell and Paul~H Radtke.
\newblock The effect of gesture on speech production and comprehension.
\newblock {\em Human factors}, 2003.

\bibitem{edwards2016jali}
Pif Edwards, Chris Landreth, Eugene Fiume, and Karan Singh.
\newblock Jali: an animator-centric viseme model for expressive lip
  synchronization.
\newblock {\em ACM Trans. on Graphics (TOG)}, 2016.

\bibitem{elhoseiny2017sherlock}
Mohamed Elhoseiny, Scott Cohen, Walter Chang, Brian Price, and Ahmed Elgammal.
\newblock Sherlock: Scalable fact learning in images.
\newblock In {\em Proc. AAAI}, 2017.

\bibitem{esser2018variational}
Patrick Esser, Ekaterina Sutter, and Bj{\"o}rn Ommer.
\newblock A variational u-net for conditional appearance and shape generation.
\newblock In {\em Proc. CVPR}, 2018.

\bibitem{flagg2009human}
Matthew Flagg, Atsushi Nakazawa, Qiushuang Zhang, Sing~Bing Kang, Young~Kee
  Ryu, Irfan Essa, and James~M Rehg.
\newblock Human video textures.
\newblock In {\em Proc. Symposium on Interactive 3D Graphics and Games}, 2009.

\bibitem{ginosar2019learning}
Shiry Ginosar, Amir Bar, Gefen Kohavi, Caroline Chan, Andrew Owens, and
  Jitendra Malik.
\newblock Learning individual styles of conversational gesture.
\newblock In {\em Proc. CVPR}, 2019.

\bibitem{gui2020featureflow}
Shurui Gui, Chaoyue Wang, Qihua Chen, and Dacheng Tao.
\newblock Featureflow: robust video interpolation via structure-to-texture
  generation.
\newblock In {\em Proc. CVPR}, 2020.

\bibitem{heck2007parametric}
Rachel Heck and Michael Gleicher.
\newblock Parametric motion graphs.
\newblock In {\em Proc. Symposium on Interactive 3D Graphics and Games}, 2007.

\bibitem{heusel2017gans}
Martin Heusel, Hubert Ramsauer, Thomas Unterthiner, Bernhard Nessler, and Sepp
  Hochreiter.
\newblock Gans trained by a two time-scale update rule converge to a local nash
  equilibrium.
\newblock In {\em Proc. NeurIPS}, 2017.

\bibitem{huang2013modeling}
Chien-Ming Huang and Bilge Mutlu.
\newblock Modeling and evaluating narrative gestures for humanlike robots.
\newblock In {\em Robotics: Science and Systems}, 2013.

\bibitem{huang2015hybrid}
Peng Huang, Margara Tejera, John Collomosse, and Adrian Hilton.
\newblock Hybrid skeletal-surface motion graphs for character animation from 4d
  performance capture.
\newblock {\em ACM Trans. on Graphics (TOG)}, 2015.

\bibitem{ilg2017flownet}
Eddy Ilg, Nikolaus Mayer, Tonmoy Saikia, Margret Keuper, Alexey Dosovitskiy,
  and Thomas Brox.
\newblock Flownet 2.0: Evolution of optical flow estimation with deep networks.
\newblock In {\em Proc. CVPR}, 2017.

\bibitem{isola2017image}
Phillip Isola, Jun-Yan Zhu, Tinghui Zhou, and Alexei~A Efros.
\newblock Image-to-image translation with conditional adversarial networks.
\newblock In {\em Proc. CVPR}, 2017.

\bibitem{iverson1998people}
Jana~M Iverson and Susan Goldin-Meadow.
\newblock Why people gesture when they speak.
\newblock {\em Nature}, 1998.

\bibitem{jiang2018super}
Huaizu Jiang, Deqing Sun, Varun Jampani, Ming-Hsuan Yang, Erik Learned-Miller,
  and Jan Kautz.
\newblock Super slomo: High quality estimation of multiple intermediate frames
  for video interpolation.
\newblock In {\em Proc. CVPR}, 2018.

\bibitem{Karras_2019_CVPR}
Tero Karras, Samuli Laine, and Timo Aila.
\newblock A style-based generator architecture for generative adversarial
  networks.
\newblock In {\em Proceedings of the IEEE/CVF Conference on Computer Vision and
  Pattern Recognition (CVPR)}, 2019.

\bibitem{kovar2008motion}
Lucas Kovar, Michael Gleicher, and Fr{\'e}d{\'e}ric Pighin.
\newblock Motion graphs.
\newblock In {\em ACM Trans. on Graphics (TOG)}. 2008.

\bibitem{kruger2010fast}
Bj{\"o}rn Kr{\"u}ger, Jochen Tautges, Andreas Weber, and Arno Zinke.
\newblock Fast local and global similarity searches in large motion capture
  databases.
\newblock In {\em Proc. ACM SCA}, 2010.

\bibitem{kucherenko2020gesticulator}
Taras Kucherenko, Patrik Jonell, Sanne van Waveren, Gustav~Eje Henter, Simon
  Alexandersson, Iolanda Leite, and Hedvig Kjellstr{\"o}m.
\newblock Gesticulator: A framework for semantically-aware speech-driven
  gesture generation.
\newblock In {\em Proc. ICMI}, 2020.

\bibitem{lee2010motion}
Yongjoon Lee, Kevin Wampler, Gilbert Bernstein, Jovan Popovi{\'c}, and Zoran
  Popovi{\'c}.
\newblock Motion fields for interactive character locomotion.
\newblock In {\em ACM Trans. on Graphics (TOG)}. 2010.

\bibitem{li2016spa}
Kun Li, Jingyu Yang, Leijie Liu, Ronan Boulic, Yu-Kun Lai, Yebin Liu, Yubin Li,
  and Eray Molla.
\newblock Spa: Sparse photorealistic animation using a single rgb-d camera.
\newblock {\em IEEE Trans. on CSVT}, 2016.

\bibitem{li2020celeb}
Yuezun Li, Xin Yang, Pu Sun, Honggang Qi, and Siwei Lyu.
\newblock Celeb-df: A large-scale challenging dataset for deepfake forensics.
\newblock In {\em Proc. CVPR}, 2020.

\bibitem{liao2020speech2video}
Miao Liao, Sibo Zhang, Peng Wang, Hao Zhu, Xinxin Zuo, and Ruigang Yang.
\newblock Speech2video synthesis with 3d skeleton regularization and expressive
  body poses.
\newblock In {\em Proc. ACCV}, 2020.

\bibitem{liu2019neural}
Lingjie Liu, Weipeng Xu, Michael Zollhoefer, Hyeongwoo Kim, Florian Bernard,
  Marc Habermann, Wenping Wang, and Christian Theobalt.
\newblock Neural rendering and reenactment of human actor videos.
\newblock {\em ACM Trans on Graphics (TOG)}, 2019.

\bibitem{liu2019liquid}
Wen Liu, Zhixin Piao, Jie Min, Wenhan Luo, Lin Ma, and Shenghua Gao.
\newblock Liquid warping gan: A unified framework for human motion imitation,
  appearance transfer and novel view synthesis.
\newblock In {\em Proc. ICCV}, 2019.

\bibitem{liu2017video}
Ziwei Liu, Raymond~A Yeh, Xiaoou Tang, Yiming Liu, and Aseem Agarwala.
\newblock Video frame synthesis using deep voxel flow.
\newblock In {\em Proc. ICCV}, 2017.

\bibitem{loper2015smpl}
Matthew Loper, Naureen Mahmood, Javier Romero, Gerard Pons-Moll, and Michael~J
  Black.
\newblock Smpl: A skinned multi-person linear model.
\newblock {\em ACM Trans. on Graphics (TOG)}, 2015.

\bibitem{mcneill1992hand}
David McNeill.
\newblock {\em Hand and mind: What gestures reveal about thought}.
\newblock University of Chicago press, 1992.

\bibitem{min2012motion}
Jianyuan Min and Jinxiang Chai.
\newblock Motion graphs++ a compact generative model for semantic motion
  analysis and synthesis.
\newblock {\em ACM Trans. on Graphics (TOG)}, 2012.

\bibitem{naert2020survey}
Lucie Naert, Caroline Larboulette, and Sylvie Gibet.
\newblock A survey on the animation of signing avatars: From sign
  representation to utterance synthesis.
\newblock {\em Computers \& Graphics}, 2020.

\bibitem{niklaus2020softmax}
Simon Niklaus and Feng Liu.
\newblock Softmax splatting for video frame interpolation.
\newblock In {\em Proc. CVPR}, 2020.

\bibitem{niklaus2017video}
Simon Niklaus, Long Mai, and Feng Liu.
\newblock Video frame interpolation via adaptive separable convolution.
\newblock In {\em Proc. ICCV}, 2017.

\bibitem{prajwal2020lip}
KR Prajwal, Rudrabha Mukhopadhyay, Vinay~P Namboodiri, and CV Jawahar.
\newblock A lip sync expert is all you need for speech to lip generation in the
  wild.
\newblock In {\em Proc. ACM International Conference on Multimedia}, 2020.

\bibitem{reitsma2007evaluating}
Paul~SA Reitsma and Nancy~S Pollard.
\newblock Evaluating motion graphs for character animation.
\newblock {\em ACM Trans. on Graphics (TOG)}, 2007.

\bibitem{ronneberger2015u}
Olaf Ronneberger, Philipp Fischer, and Thomas Brox.
\newblock U-net: Convolutional networks for biomedical image segmentation.
\newblock In {\em International Conference on Medical image computing and
  computer-assisted intervention}, 2015.

\bibitem{rossler2019faceforensics++}
Andreas Rossler, Davide Cozzolino, Luisa Verdoliva, Christian Riess, Justus
  Thies, and Matthias Nie{\ss}ner.
\newblock Faceforensics++: Learning to detect manipulated facial images.
\newblock In {\em Proc. ICCV}, 2019.

\bibitem{rubin1977locus}
Steven~M Rubin and Raj Reddy.
\newblock The locus model of search and its use in image interpretation.
\newblock In {\em IJCAI}, 1977.

\bibitem{safonova2007construction}
Alla Safonova and Jessica~K Hodgins.
\newblock Construction and optimal search of interpolated motion graphs.
\newblock {\em ACM Trans. on Graphics (TOG)}, 2007.

\bibitem{schodl2000video}
Arno Sch{\"o}dl, Richard Szeliski, David~H Salesin, and Irfan Essa.
\newblock Video textures.
\newblock In {\em Proc. Conference on Computer Graphics and Interactive
  Techniques}, 2000.

\bibitem{seshadrinathan2009motion}
Kalpana Seshadrinathan and Alan~Conrad Bovik.
\newblock Motion tuned spatio-temporal quality assessment of natural videos.
\newblock {\em IEEE Trans. Image Processing}, 2009.

\bibitem{shin2006fat}
Hyun~Joon Shin and Hyun~Seok Oh.
\newblock Fat graphs: constructing an interactive character with continuous
  controls.
\newblock In {\em Proc. ACM SCA}, 2006.

\bibitem{shum2001construction}
H-Y Shum and Richard Szeliski.
\newblock Construction of panoramic image mosaics with global and local
  alignment.
\newblock In {\em Panoramic vision}. 2001.

\bibitem{shysheya2019textured}
Aliaksandra Shysheya, Egor Zakharov, Kara-Ali Aliev, Renat Bashirov, Egor
  Burkov, Karim Iskakov, Aleksei Ivakhnenko, Yury Malkov, Igor Pasechnik,
  Dmitry Ulyanov, et~al.
\newblock Textured neural avatars.
\newblock In {\em Proc. CVPR}, 2019.

\bibitem{siarohin2018deformable}
Aliaksandr Siarohin, Enver Sangineto, St{\'e}phane Lathuiliere, and Nicu Sebe.
\newblock Deformable gans for pose-based human image generation.
\newblock In {\em Proc. CVPR}, 2018.

\bibitem{siarohin2021motion}
Aliaksandr Siarohin, Oliver~J Woodford, Jian Ren, Menglei Chai, and Sergey
  Tulyakov.
\newblock Motion representations for articulated animation.
\newblock In {\em Proc. CVPR}, 2021.

\bibitem{simonyan2014very}
Karen Simonyan and Andrew Zisserman.
\newblock Very deep convolutional networks for large-scale image recognition.
\newblock {\em ICLR}, 2015.

\bibitem{taylor2017deep}
Sarah Taylor, Taehwan Kim, Yisong Yue, Moshe Mahler, James Krahe,
  Anastasio~Garcia Rodriguez, Jessica Hodgins, and Iain Matthews.
\newblock A deep learning approach for generalized speech animation.
\newblock {\em ACM Trans. on Graphics (TOG)}, 2017.

\bibitem{teed2020raft}
Zachary Teed and Jia Deng.
\newblock Raft: Recurrent all-pairs field transforms for optical flow.
\newblock In {\em Proc. ECCV}, 2020.

\bibitem{thies2020neural}
Justus Thies, Mohamed Elgharib, Ayush Tewari, Christian Theobalt, and Matthias
  Nie{\ss}ner.
\newblock Neural voice puppetry: Audio-driven facial reenactment.
\newblock In {\em Proc. ECCV}, 2020.

\bibitem{wang2004seamless}
Hongcheng Wang, Ramesh Raskar, and Narendra Ahuja.
\newblock Seamless video editing.
\newblock In {\em Proc. ICPR}, 2004.

\bibitem{kaisiyuan2020mead}
Kaisiyuan Wang, Qianyi Wu, Linsen Song, Zhuoqian Yang, Wayne Wu, Chen Qian, Ran
  He, Yu Qiao, and Chen~Change Loy.
\newblock Mead: A large-scale audio-visual dataset for emotional talking-face
  generation.
\newblock In {\em Proc. ECCV}, 2020.

\bibitem{wang2019few}
Ting-Chun Wang, Ming-Yu Liu, Andrew Tao, Guilin Liu, Bryan Catanzaro, and Jan
  Kautz.
\newblock Few-shot video-to-video synthesis.
\newblock In {\em Proc. NeurIPS}, 2019.

\bibitem{wang2018video}
Ting-Chun Wang, Ming-Yu Liu, Jun-Yan Zhu, Guilin Liu, Andrew Tao, Jan Kautz,
  and Bryan Catanzaro.
\newblock Video-to-video synthesis.
\newblock In {\em Proc. NeurIPS}, 2018.

\bibitem{wang2018high}
Ting-Chun Wang, Ming-Yu Liu, Jun-Yan Zhu, Andrew Tao, Jan Kautz, and Bryan
  Catanzaro.
\newblock High-resolution image synthesis and semantic manipulation with
  conditional gans.
\newblock In {\em Proc. CVPR}, 2018.

\bibitem{weng2019photo}
Chung-Yi Weng, Brian Curless, and Ira Kemelmacher-Shlizerman.
\newblock Photo wake-up: 3d character animation from a single photo.
\newblock In {\em Pro. CVPR}, 2019.

\bibitem{weng2020vid2actor}
Chung-Yi Weng, Brian Curless, and Ira Kemelmacher-Shlizerman.
\newblock Vid2actor: Free-viewpoint animatable person synthesis from video in
  the wild.
\newblock {\em arXiv preprint arXiv:2012.12884}, 2020.

\bibitem{xiang2019monocular}
Donglai Xiang, Hanbyul Joo, and Yaser Sheikh.
\newblock Monocular total capture: Posing face, body, and hands in the wild.
\newblock In {\em Proc. CVPR}, 2019.

\bibitem{xiong2018microsoft}
Wayne Xiong, Lingfeng Wu, Fil Alleva, Jasha Droppo, Xuedong Huang, and Andreas
  Stolcke.
\newblock The microsoft 2017 conversational speech recognition system.
\newblock In {\em Proc. ICASSP}, 2018.

\bibitem{xu2011video}
Feng Xu, Yebin Liu, Carsten Stoll, James Tompkin, Gaurav Bharaj, Qionghai Dai,
  Hans-Peter Seidel, Jan Kautz, and Christian Theobalt.
\newblock Video-based characters: creating new human performances from a
  multi-view video database.
\newblock In {\em ACM Trans. on Graphics (TOG)}. 2011.

\bibitem{yang2020statistics}
Yanzhe Yang, Jimei Yang, and Jessica Hodgins.
\newblock Statistics-based motion synthesis for social conversations.
\newblock In {\em Computer Graphics Forum}, 2020.

\bibitem{yoon2020speech}
Youngwoo Yoon, Bok Cha, Joo-Haeng Lee, Minsu Jang, Jaeyeon Lee, Jaehong Kim,
  and Geehyuk Lee.
\newblock Speech gesture generation from the trimodal context of text, audio,
  and speaker identity.
\newblock {\em ACM Trans. on Graphics (TOG)}, 2020.

\bibitem{yunus2020sequence}
Fajrian Yunus, Chlo{\'e} Clavel, and Catherine Pelachaud.
\newblock Sequence-to-sequence predictive models: from prosody to communicative
  gestures.
\newblock In {\em Workshop sur les Affects, Compagnons artificiels et
  Interactions}, 2020.

\bibitem{zakharov2019few}
Egor Zakharov, Aliaksandra Shysheya, Egor Burkov, and Victor Lempitsky.
\newblock Few-shot adversarial learning of realistic neural talking head
  models.
\newblock In {\em Proc. ICCV}, 2019.

\bibitem{zhang2020vid2player}
Haotian Zhang, Cristobal Sciutto, Maneesh Agrawala, and Kayvon Fatahalian.
\newblock Vid2player: Controllable video sprites that behave and appear like
  professional tennis players.
\newblock {\em arXiv preprint arXiv:2008.04524}, 2020.

\bibitem{zhang2018unreasonable}
Richard Zhang, Phillip Isola, Alexei~A Efros, Eli Shechtman, and Oliver Wang.
\newblock The unreasonable effectiveness of deep features as a perceptual
  metric.
\newblock In {\em Proc. CVPR}, 2018.

\bibitem{zheng2019unsupervised}
Haitian Zheng, Lele Chen, Chenliang Xu, and Jiebo Luo.
\newblock Unsupervised pose flow learning for pose guided synthesis.
\newblock {\em arXiv}, 2019.

\bibitem{zhou2019talking}
Hang Zhou, Yu Liu, Ziwei Liu, Ping Luo, and Xiaogang Wang.
\newblock Talking face generation by adversarially disentangled audio-visual
  representation.
\newblock In {\em Proc. AAAI}, 2019.

\bibitem{zhou2021face}
Tianfei Zhou, Wenguan Wang, Zhiyuan Liang, and Jianbing Shen.
\newblock Face forensics in the wild.
\newblock In {\em Proc. CVPR}, 2021.

\bibitem{zhou2020makelttalk}
Yang Zhou, Xintong Han, Eli Shechtman, Jose Echevarria, Evangelos Kalogerakis,
  and Dingzeyu Li.
\newblock Makelttalk: speaker-aware talking-head animation.
\newblock {\em ACM Trans. on Graphics (TOG)}, 2020.

\bibitem{zhou2018visemenet}
Yang Zhou, Zhan Xu, Chris Landreth, Evangelos Kalogerakis, Subhransu Maji, and
  Karan Singh.
\newblock Visemenet: Audio-driven animator-centric speech animation.
\newblock {\em ACM Trans. on Graphics (TOG)}, 2018.

\end{thebibliography}
}

\clearpage
\section*{Supplementary Materials}
\paragraph{Implementation and demo videos.} Our implementation and demo videos can be found at
our project page \mbox{\url{https://yzhou359.github.io/video_reenact}}.

\paragraph{Training details.}
We train the entire network end-to-end with losses promoting better flow estimation and final frame reconstruction. Specifically, we first have an L1 reconstruction loss $L_{rec}$ and a perceptual loss $L_{per}$ between the synthesized image $\hat{I}_t$ and $I_t$:

\begin{align}
    L_{rec} &= \L_1 (I_t, \hat{I}_t) \\
    L_{per} &= \L_1 (\phi(I_t), \phi(\hat{I}_t))
\end{align}
where $\phi(\cdot)$ concatenates feature map activations from a pre-trained VGG19 network~\cite{simonyan2014very}.

We then adopt another L1 reconstruction loss $\L^{b}_{rec}$ promoting better frame reconstruction directly from  the warped deep features $x''_i$ and $x''_j$ after these pass through our generator network $G$. This helped predict warped deep features such that they lead to generating frames as close as possible to ground-truth in the first place.
We also empirically observed faster convergence with this loss:

\begin{equation}
    L^{b}_{rec} = \L_1 (I_t, G(x''_i)) + \L_1 (I_t, G(x''_j))
\end{equation}

Further, we have warping loss $L^m_{warp}$ and $L^o_{warp}$ by measuring the L1 reconstruction error between the target image and the source images $I_i$ and $I_j$ after being warped through the motion field $F^m_{t\to i}$ (Equations 2 and 3 in the main paper) and also the optical flow $F^o_{t\to i}$:

\begin{align}
\begin{split}
    L^{m}_{warp} &= \L_1 (I_t, \W(I_i, F^m_{t\to i})) + \\
    &\ \ \ \ \ \ \ \ \ \ \L_1 (I_t, \W(I_j, F^m_{t\to j})) \\
\end{split}
\end{align}
\begin{align}
\begin{split}
     L^{o}_{warp} &= \L_1 (I_t, \W(\W(I_i, F^m_{t\to i}), F^o_{t\to i})) + \\
    &\ \ \ \ \ \ \ \ \ \ \L_1 (I_t, \W(\W(I_j, F^m_{t\to j}), F^o_{t\to j}))
\end{split}
\end{align}
where $\W(I, F)$ applies backward warping flow $F$ on image $I$.

\begin{table}[t]
\small
    \centering
    \begin{tabular}{c|l}
     Category  & Keywords  \\
     \hline \hline
     greeting  & hey, hi, hello \\ \hline
     counting  & one, two, three, first, second, third \\ \hline
     direction & east, west, north, south, back, front, away, \\
     & here, around \\ \hline
     sentiment & crazy, incredible, surprising, screaming \\ \hline
     action & walk, drive, ride, enter, open, attach, take, move \\ \hline
     relative & more, less, much, few \\
     \hline
     others & called
    \end{tabular}
    \vspace{1mm}
    \caption{Dictionary of common keywords.}
    \label{tab:dic}
\end{table}

\begingroup
\setlength{\tabcolsep}{1pt} 
\begin{table*}[ht]
\begin{center}
\begin{tabular}{c c c c c c c c c}
\includegraphics[width=0.1\linewidth]{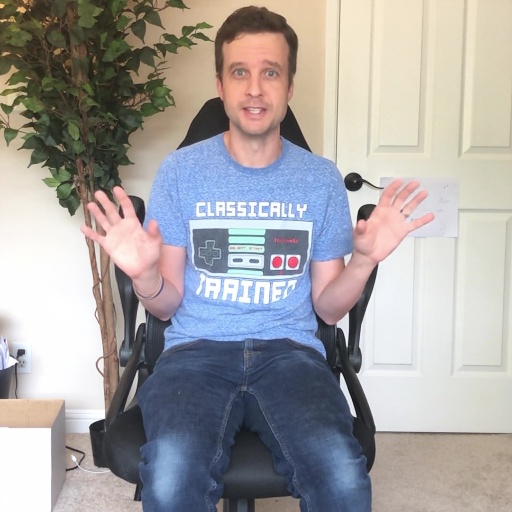} &
\includegraphics[width=0.1\linewidth]{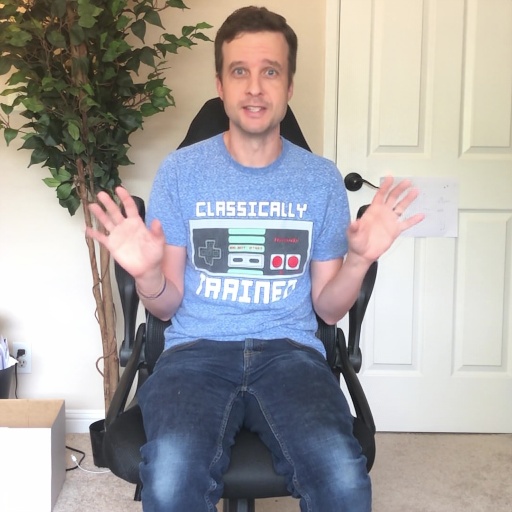} &
\includegraphics[width=0.1\linewidth]{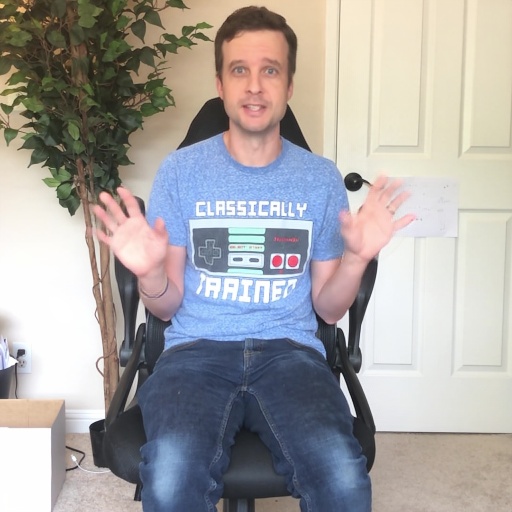} &
\includegraphics[width=0.1\linewidth]{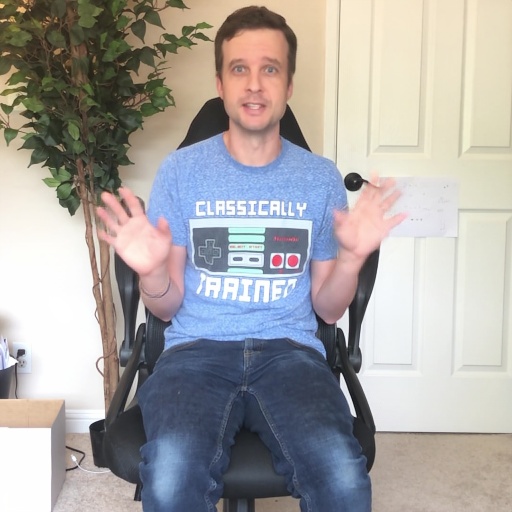} &
\includegraphics[width=0.1\linewidth]{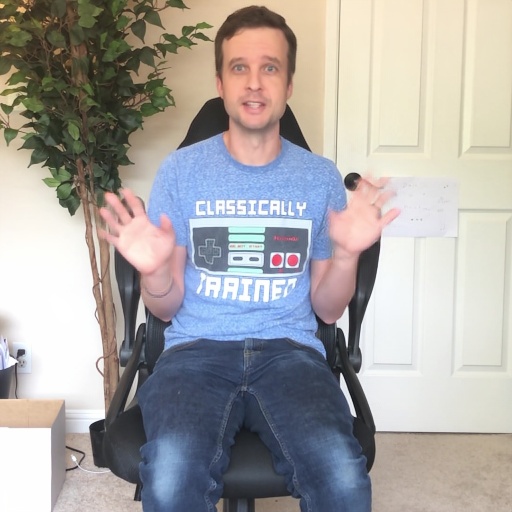} &
\includegraphics[width=0.1\linewidth]{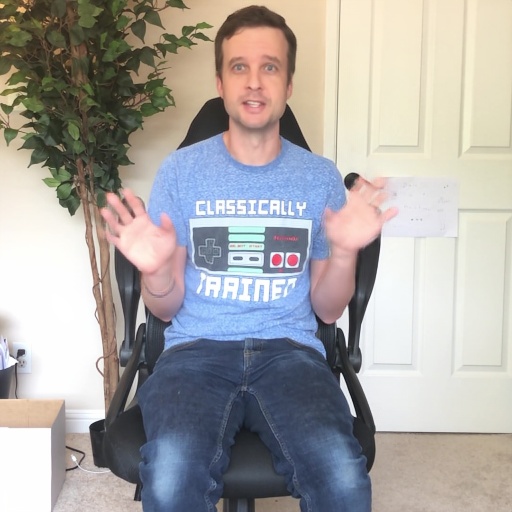} &
\includegraphics[width=0.1\linewidth]{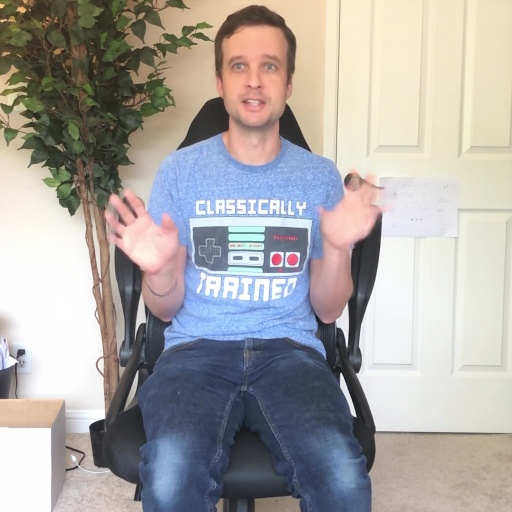} &
\includegraphics[width=0.1\linewidth]{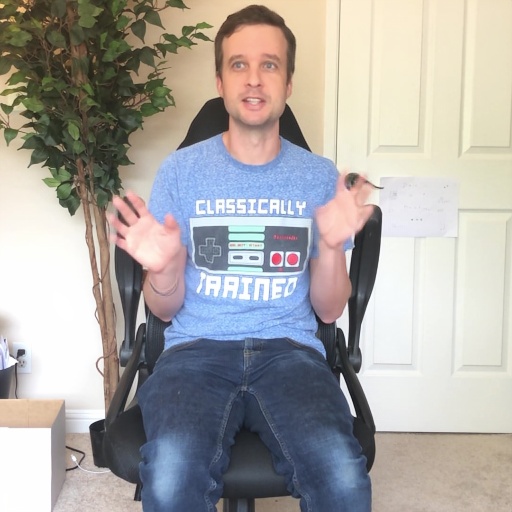} &
\includegraphics[width=0.1\linewidth]{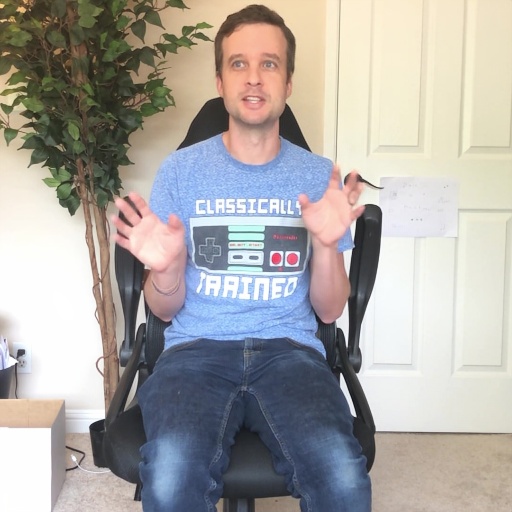} \\

$I_i$ & $\alpha=1/8$ & $\alpha=1/4$ & $\alpha=3/8$ & $\alpha=1/2$ & $\alpha=5/8$ & $\alpha=3/4$ & $\alpha=7/8$ & $I_j$\\
\end{tabular}
\begin{tabular}{c c c c c c c c c c}
$F^m_{t\to i}$ &
\includegraphics[width=0.09\linewidth]{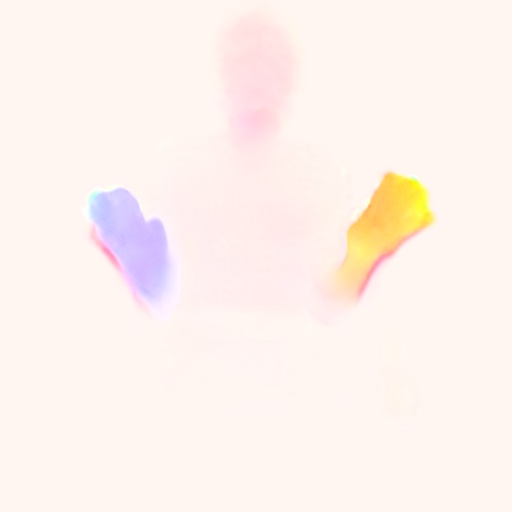} &
\includegraphics[width=0.09\linewidth]{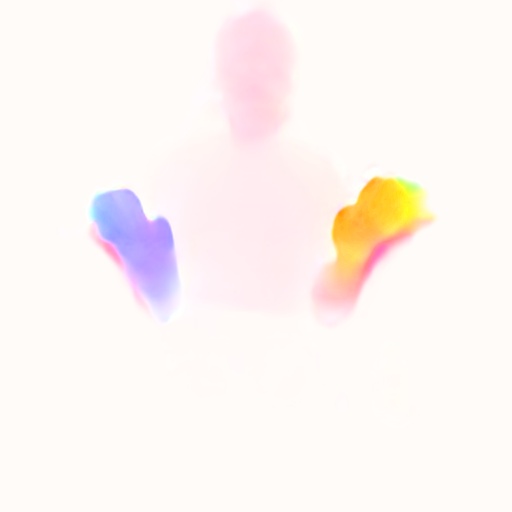} &
\includegraphics[width=0.09\linewidth]{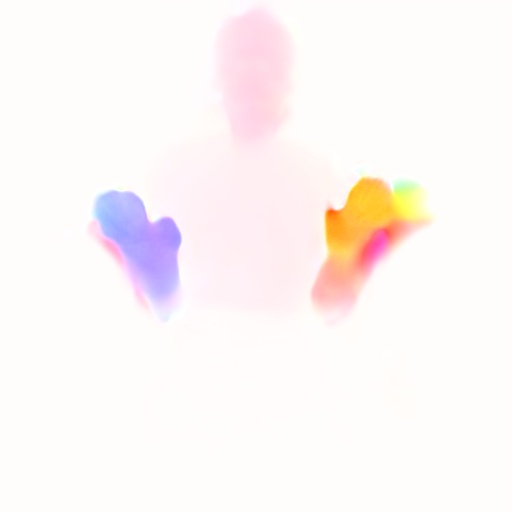} &
\includegraphics[width=0.09\linewidth]{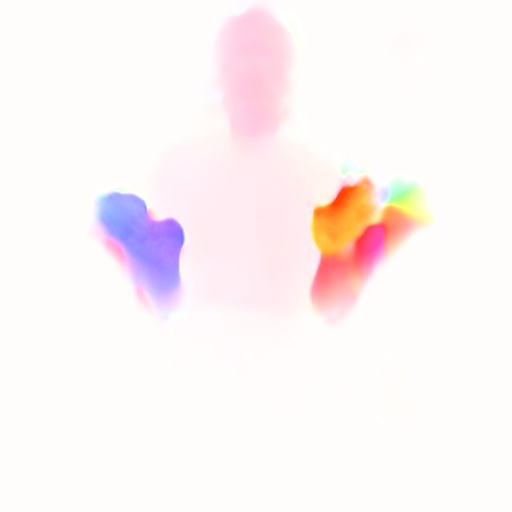} &
$F^m_{t\to j}$ &
\includegraphics[width=0.09\linewidth]{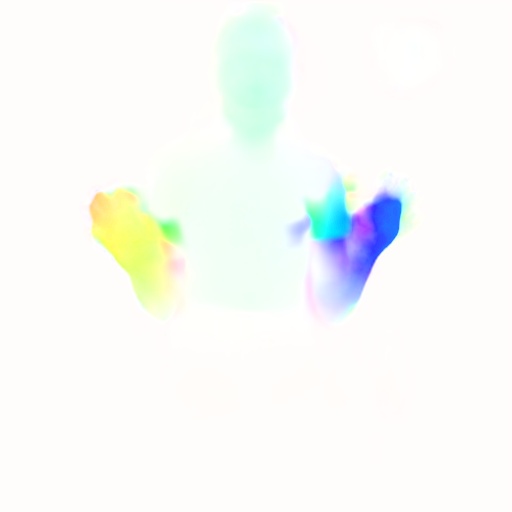} &
\includegraphics[width=0.09\linewidth]{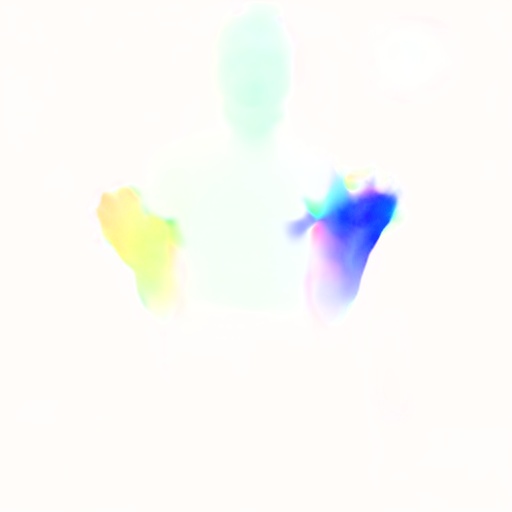} &
\includegraphics[width=0.09\linewidth]{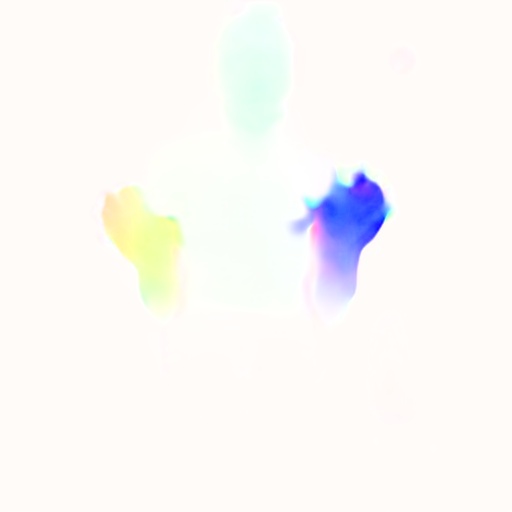} &
\includegraphics[width=0.09\linewidth]{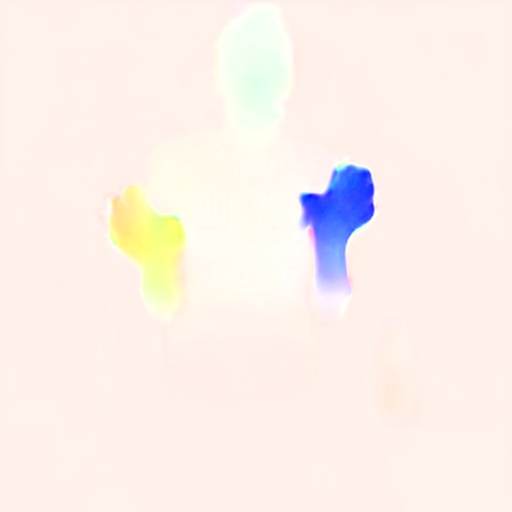}
 \\
 
$F^o_{t\to i}$ &
\includegraphics[width=0.09\linewidth]{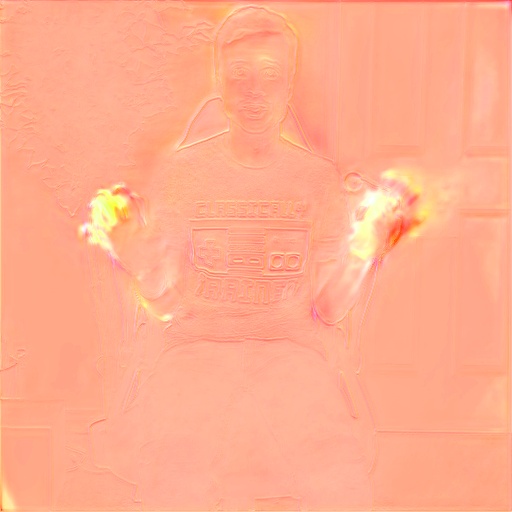} &
\includegraphics[width=0.09\linewidth]{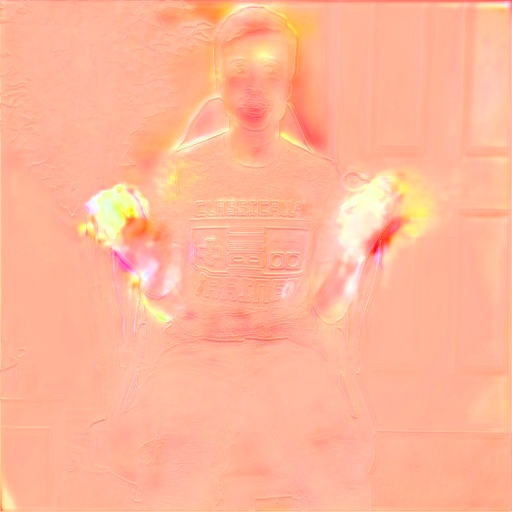} &
\includegraphics[width=0.09\linewidth]{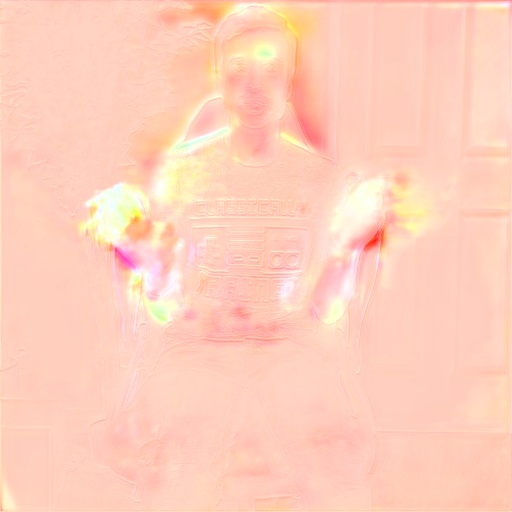} &
\includegraphics[width=0.09\linewidth]{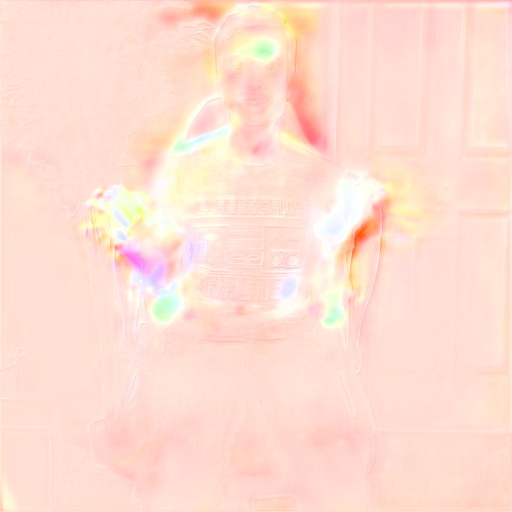} &
$F^o_{t\to j}$ &
\includegraphics[width=0.09\linewidth]{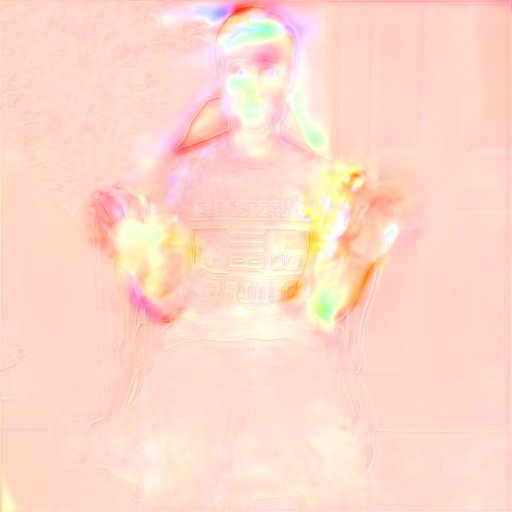} &
\includegraphics[width=0.09\linewidth]{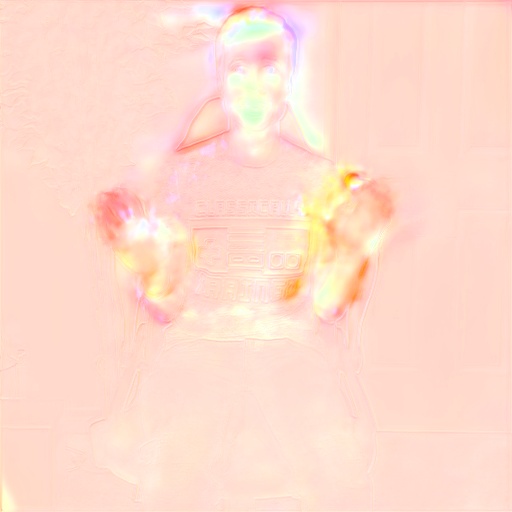} &
\includegraphics[width=0.09\linewidth]{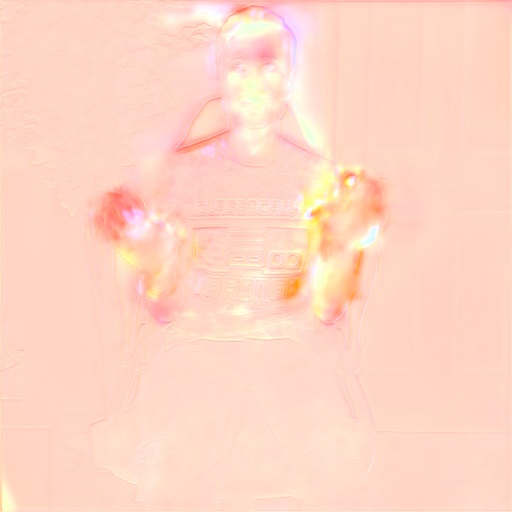} &
\includegraphics[width=0.09\linewidth]{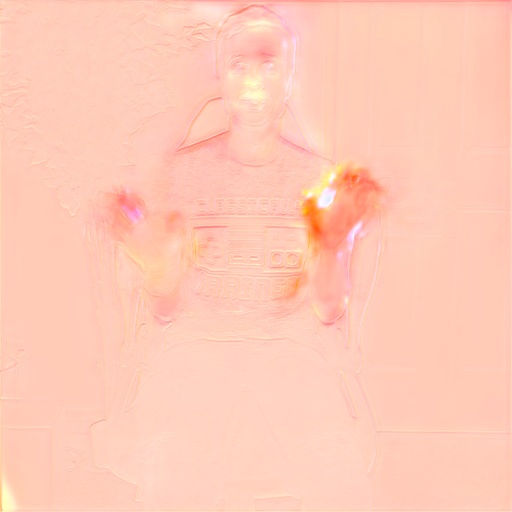}
 \\
 
$(1-\alpha)V_{t\to i}\ \ $ &
\includegraphics[width=0.09\linewidth]{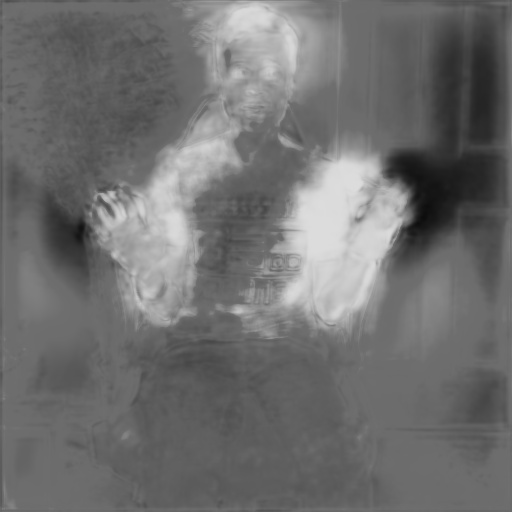} &
\includegraphics[width=0.09\linewidth]{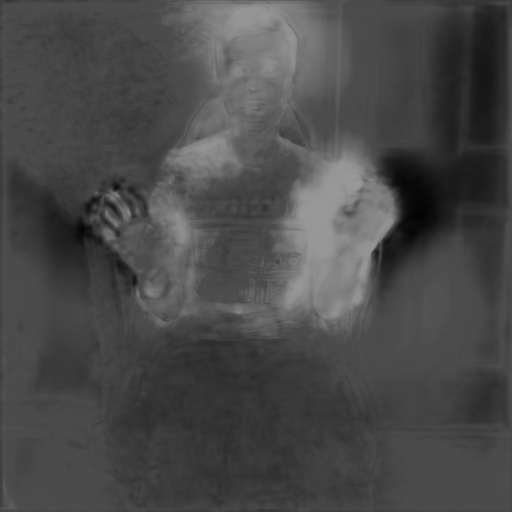} &
\includegraphics[width=0.09\linewidth]{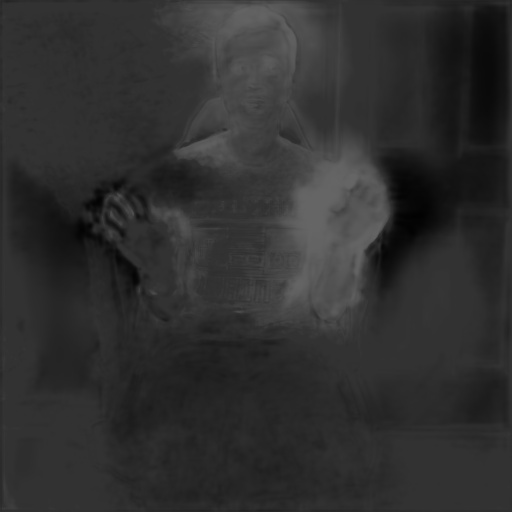} &
\includegraphics[width=0.09\linewidth]{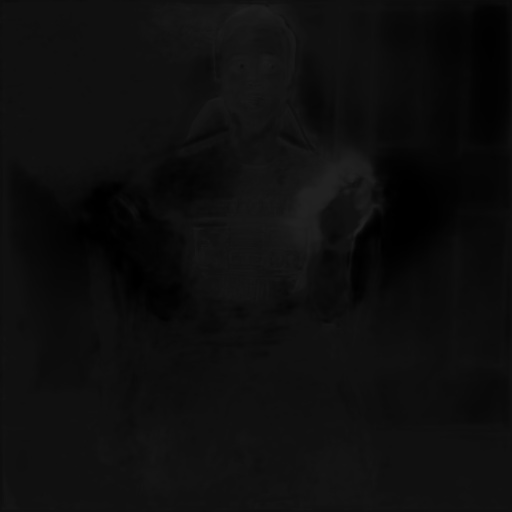} &
$\ \ \alpha V_{t\to j}\ \ $ &
\includegraphics[width=0.09\linewidth]{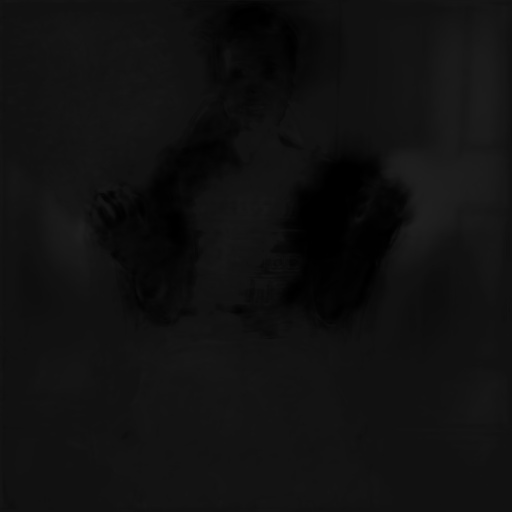} &
\includegraphics[width=0.09\linewidth]{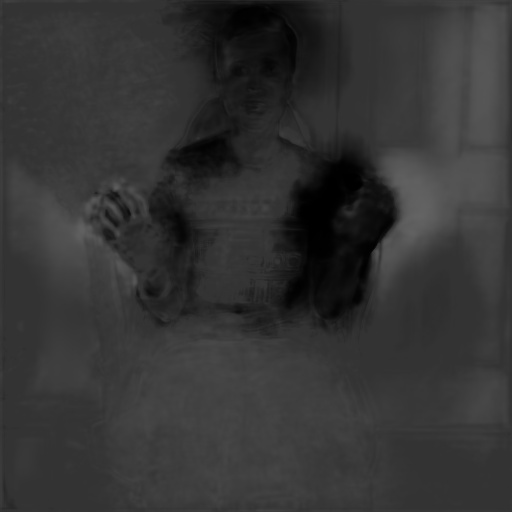} &
\includegraphics[width=0.09\linewidth]{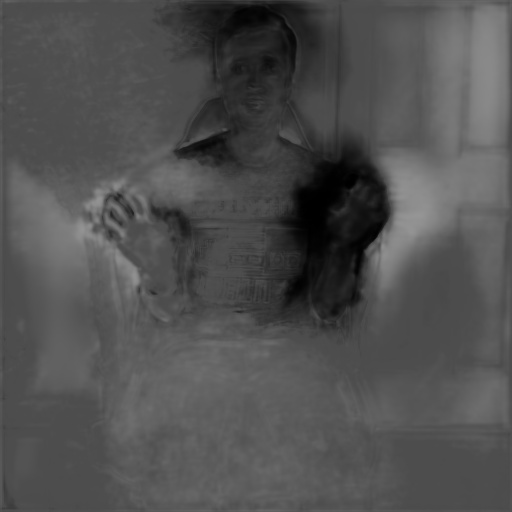} &
\includegraphics[width=0.09\linewidth]{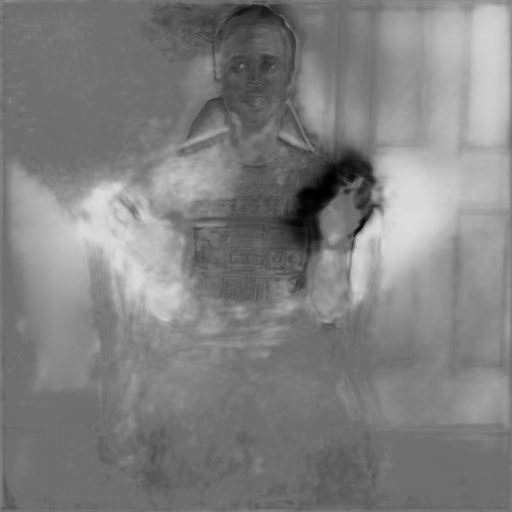}
 \\
 
 & $\alpha=1/8$ & $\alpha=3/8$ & $\alpha=5/8$ & $\alpha=7/8$ & & $\alpha=1/8$ & $\alpha=3/8$ & $\alpha=5/8$ & $\alpha=7/8$
\end{tabular}
\end{center}
\captionof{figure}{Pose-aware video blending results for target blending weights $\alpha \in (0, 1)$. \textbf{Top row}: synthesized in-between frames with blended human gestures for different blending weights. \textbf{Bottom rows}: intermediate mesh flows, optical flows and visibility maps results for corresponding blending weights.}
\label{fig:ours_interpolate_t}
\end{table*}
\endgroup

Finally, we follow \cite{jiang2018super} and include a smoothness loss for both mesh flow and optical flow: 

\begin{align}
    L_{sm} = ||\nabla &F^{m}_{t\to i}||_1 + ||\nabla F^{m}_{t\to j}||_1 +\\
    &||\nabla F^{o}_{t\to i}||_1 + ||\nabla F^{o}_{t\to j}||_1
\end{align}

The overall loss $\L$ is defined as the weighted sum of all losses described above, then averaged over all training frames.

\begin{align}
\begin{split}
    \L = \L_{rec} + \lambda_{p}& \L_{per} + \lambda_b \L^b_{rec} + \\
    &\lambda_m \L^m_{warp} + \lambda_o \L^o_{warp} + \lambda_s \L_{sm}
\end{split}
\end{align}
The weights have been set empirically based on \cite{jiang2018super}
as $\lambda_{p}=0.01,\ \lambda_b=0.25,\ \lambda_m=0.25,\ \lambda_o=0.25, \lambda_s=0.01$.

To train the entire model, we first train the mesh flow estimator network with $L^{m}_{warp}$ as a ``warming'' stage. Then we load a pre-trained optical flow model from \cite{jiang2018super}. Finally, we train the entire network end-to-end with the loss mentioned above. The network weights are optimized with Adam optimizer using PyTorch. The learning rate is set to $10^{-4}$ and weight decay to $10^{-6}$. The training process is performed on 4 Nvidia GeForce 1080Ti GPUs.

We show the detailed training procedure for our numerical evaluation. For the Personal story dataset, we train the model on each individual speaker video and report the evaluation numbers accordingly. The compared methods are trained on each speaker video for a fair comparison. For the TED-talks dataset, we train a single model on the entire training split of the dataset. We evaluate our model generalization on the testing split which contains unseen speakers. The comparison methods are also trained on multiple speakers on this dataset for a fair comparison.

The TED-Talks dataset proposed by \cite{siarohin2021motion} contains a list of Youtube video URL links, corresponding frame indices, and cropped areas with auto-detected upper bodies of speakers inside. They are not directly helpful to create the audio-driven reenacted video results. This is because 1) the original dataset only contains very short video clips, e.g., with a duration of a few seconds, which are not sufficient to create rich video motion graphs; 2) the frames in the original dataset are processed to $384 \times 384$ resolution by cropping and scaling the upper body of the speaker from a zoomed-out full body frame. As a result, the frames do not have high resolution and high quality. In this case, we use such dataset for numerical evaluation and easier reproduction purpose. To achieve high resolution and high quality audio-driven reenacted TED-talks videos shown on our project page, we use the original full Youtube videos. We manually select the frames with the zoomed-in camera where the upper body of the speaker appears at high resolution and high quality (see examples in our HTML files). The selected frames have sufficient length to create reasonable video motion graphs. Finally we fine-tune the model on these frames from each specific speaker and generate reenacted video results given test audios.

\paragraph{Pose-aware video blending network results.}
Fig.~\ref{fig:ours_interpolate_t} shows output images from the
video blending network  for different blending weights, along with 
results from our intermediate stages.

\paragraph{Dictionary of common keywords.}

Referential gestures, especially iconic and metaphoric gestures, have strong correlations with the transcript~\cite{mcneill1992hand,yoon2020speech}. They usually appear together with certain keywords, such as action verbs, concrete objects, abstract concepts, and relative quantities to co-express the speech content~\cite{huang2013modeling}. We gather a few frequently used such keywords co-occurring with referential gestures in our speaker videos, as shown in Table.~\ref{tab:dic}.

\paragraph{Network architecture details.}

The spatial encoder network $E_s$ takes as input the RGB image $I_i$, the foreground mask $I_{mask}$, and an image containing the rendered skeleton $I_{skel}$ representing the SMPL pose parameters. Fig.~\ref{fig:encode_input} shows an example of these input images.


We show our \textit{Spatial Encoder} network structure for generating the mesh flow warping field in Table~\ref{tab:mesh_flow}. In this table, the left column indicates the spatial resolution of the feature map output.  The
\textit{ResBlock down} block is a 2-strided convolutional layer with a $3\times3$ kernel followed by two residual blocks. The \textit{ResBlock up} block is a nearest-neighbor upsampling with a scale of 2, followed by a a $3\times3$ convolutional layer and then two residual blocks. The term \textit{Skip} means skip connection that concatenates the feature maps of an encoding layer and decoding layer with the same spatial resolution. For Personal story dataset, the input and generated images are in $512 \times 512$ resolution, while for TED-talks dataset, the image resolution is $384 \times 384$.

The \textit{Mesh Flow Estimator} and \textit{Image Generator} network follows the structure of the \textit{Spatial Encoder} network (see Table~\ref{tab:mesh_flow}), but the input and output number of channels are different. For the \textit{Mesh Flow Estimator} network, the number of input feature channel is 13 and output feature channel is 2. For the \textit{Image Generator} network, the number of input feature channel is 19 and output feature channel is 3. Besides, the \textit{Image Generator} network uses in the end a $tanh(\cdot)$ activation  to regularize the image values between $[0, 1]$.

\revise{
\paragraph{Audio-driven Beam search details.}
We initialize a  beam search \cite{rubin1977locus} procedure in the video motion graph to find $K$ plausible paths matching the target speech audio segments. We set K to 20.
The beam search initializes $K$ paths starting with $K$ random nodes as the first frame $a_0$ for the target audio, then expands in a breadth-first-search manner to find paths ending at a \emph{target graph node} whose audio feature matches the target audio feature at the endpoint of the first segment $a_1$, associated with either an activated audio onset or the same non-empty keyword feature.
Note that there can be multiple target graph nodes sharing the same audio feature with  $a_1$.

During the beam search, all the explored paths are sorted based on a \textit{path transition cost}, plus a \textit{path duration cost}.
The path transition cost is defined as the sum of node distances between all consecutive nodes $m, n$ along the path, i.e. $\sum_{m,n} {(d_{feat}(m,n) + d_{img}(m,n))}$. 
The cost of synthetic transitions are always higher than natural ones. Thus, the path cost prevents using implausible paths with too many synthetic transitions.

When a path reaches a \emph{target} graph node, we check its duration. 
Due to the sparsity of the graph, there may not be any path matching exactly the target audio segment length $L_i$.
Still, the path length should be similar to $L_i$, otherwise one would need to accelerate or decelerate the path too much to adjust it to the exact length, leading to unnaturally fast or slow gestures. 
We only accept paths with
duration $L'_s \in [0.9L_s,1.1L_s]$ since these can be slightly adjusted, e.g. re-sampled,  to match the target segment duration.
For the above range, we observed that the motion still looks natural. 
Nevertheless, we also add a path duration cost $|1 - L'_s/L_s|$ to favor paths during beam search with duration closer to the target duration.  

\rev{When the speech audio is silent, the searched motion graph paths go through nodes without audio onset features, which are often the frames with rest poses.}

After processing the first segment $a_0 \to a_{1}$, we start another beam search for the next segment $a_1 \to a_2$.  
Here, the path expansion starts with the last node of the $K$ paths discovered from previous iteration. 
The expansion continues with the same search procedure as above.
In order, the searches run iteratively for all the rest segments $a_s \to a_{s+1}$, $s\in[1, S]$ while always keeping the most plausible $K$ paths. 
All searched $K$ paths can be used to generate various plausible results for the same target speech audio (see demo videos on our project page). 
The best path is picked in our experiments.
}

\begingroup
\setlength{\tabcolsep}{1pt} 
\begin{figure}
    \centering
    \begin{tabular}{c  c  c}
    \includegraphics[width=0.16\textwidth]{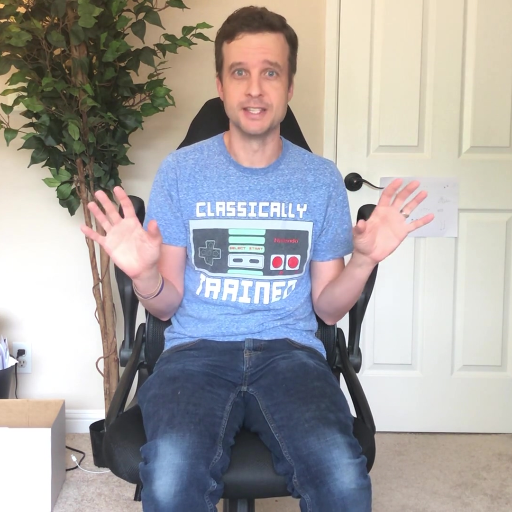} &
    \includegraphics[width=0.16\textwidth]{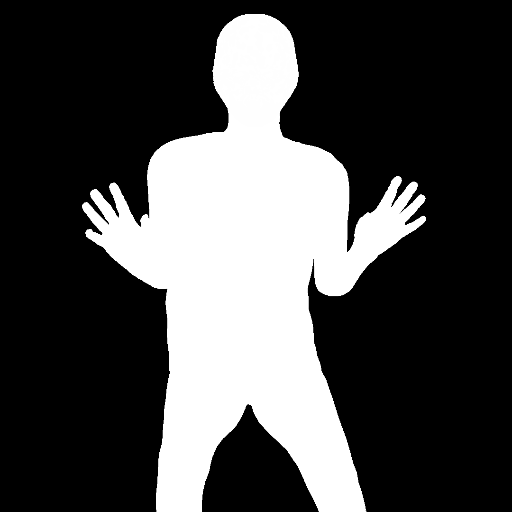} &
    \includegraphics[width=0.16\textwidth]{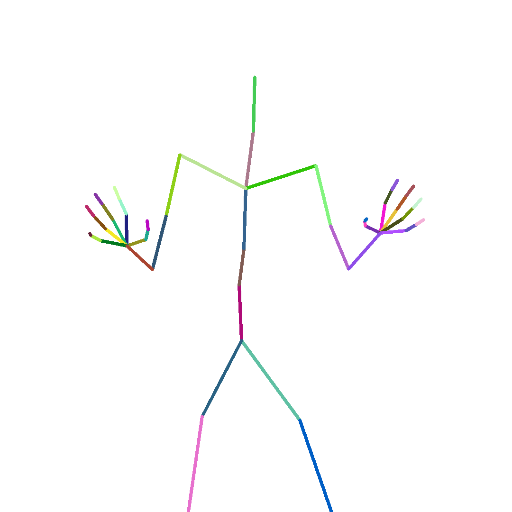} \\
    (a) & (b) & (c)
    \end{tabular}
    \caption{An example of inputs to our spatial encoder network (a) input image frame,  (b) corresponding foreground human mask and  (c) rendered skeleton image.}
    \label{fig:encode_input}
\end{figure}
\endgroup

\paragraph{User study details.}
We provide here more details about the user study. 

\begin{figure}[t!]
    \centering
    \includegraphics[width=\linewidth]{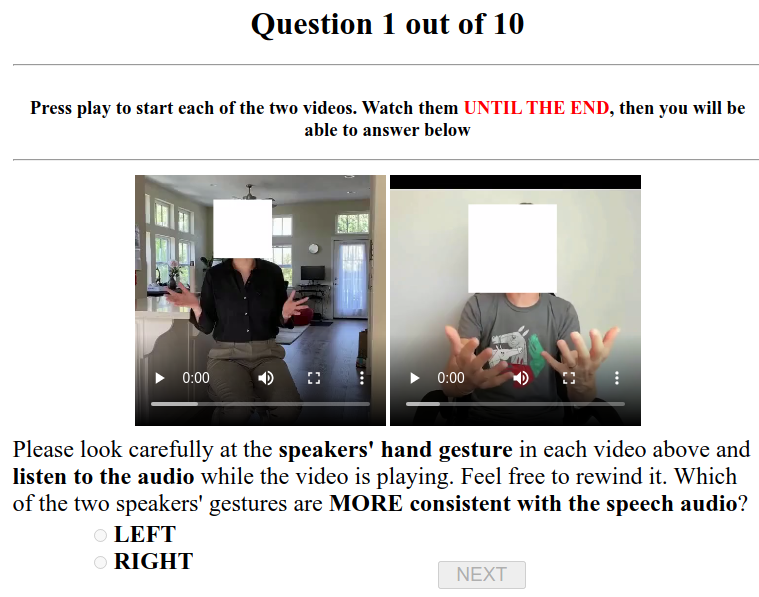}
    \caption{User study questionnaire page.}
    \label{fig:user_study_page}
\end{figure}

We have a pool of 381 queries (127 videos from each method $\times$ 3 comparison pairs).
For each query, we show two videos in parallel randomly placed at left/right positions. The participants are asked which speaker's gestures are more consistent with the speech audio and vote for one of the two choices: “left animation”, “right animation”.
Fig.~\ref{fig:user_study_page} shows the webpage layout used in our questionnaires. The layout shows two video results to the participants, a  question on the bottom and two choices (``left''/``right''). To enable the selection of either choice, the users must watch both videos until the end. We also explicitly instruct them to focus on the speakers' hand gestures and ignore the masked facial area.

Our questionnaires also include a similar page layout showing tutorial examples in the beginning.  The tutorial shows a pair of videos with clear differences: one video is from ground-truth in which the speaker's gestures are naturally consistent with the audio; the other video is a failure case, which shows gestures that are inconsistent with audio at some places. For these tutorial cases only, we let the participants pick an answer first and then let them know whether their answer is correct or wrong and explain why.

We also adapt a user validation check to filter out unreliable MTurkers. Specifically, after the tutorial,
our questionnaires showed $10$ queries in a random order. $3$ of the  queries were repeated twice (i.e., we had $7$ unique queries per questionnaire). We randomly flipped the two videos each time to detect unreliable participants giving inconsistent answers. We filter out unreliable MTurk participants who give different answers to two (or more) of the repeated queries in the questionnaire or took less than $5$ minutes to complete it. 
Each participant was allowed to answer one questionnaire maximum to ensure participant diversity. 
We collected answers from 
$113$ reliable participants for our user study.
\rev{We paid \$1 per questionnaire. All comparison outcomes are statistically significant using a z-test (p$<$.05).}

\begin{table}
\centering
\begin{tabular}{c|c}
\hline
\multirow{2}{*}{$512 \times 512$}&Input RGB image, foreground mask image, \\
&and rendered skeleton image\\  
\hline
$256\times256$ & ResBlock down $(16+2)\rightarrow 32$ \\
\hline
$128\times128$ & ResBlock down $32\rightarrow 64$ \\
\hline
$64\times64$ & ResBlock down $64\rightarrow 128$ \\
\hline
$32\times32$ & ResBlock down $128\rightarrow 256$ \\
\hline
$16\times16$ & ResBlock down $256\rightarrow 512$ \\
\hline
$8\times8$ & ResBlock down $512\rightarrow 512$ \\
\hline
$8\times8$ & ResBlock up $512 \rightarrow 512$ \\
\hline
$16\times16$ & Skip + ResBlock up $(512+512)\rightarrow 512$ \\
\hline
$32\times32$ & Skip + ResBlock up $(512+512)\rightarrow 256$ \\
\hline
$64\times64$ & Skip + ResBlock up $(256+256)\rightarrow 128$ \\
\hline
$128\times128$ & Skip + ResBlock up $(128+128)\rightarrow 64$ \\
\hline
$256\times256$ & Skip + ResBlock up $(64+64)\rightarrow 32$ \\
\hline
$512 \times 512$ & Skip + ResBlock up $(32+32)\rightarrow 16$ \\
\hline
\end{tabular}
\caption{Spatial Encoder network structure.}
\label{tab:mesh_flow}
\end{table}

\rev{\paragraph{Importance of the reference video.} The key idea of using reference video is that it provides personalized gestures. Directly animating a single portrait is hard since it is not clear what are the `\textit{correct}' gestures. There are many applications of our setup. For example, in video production, there is a need to add or remove sentences from existing clips. In online education, different video lessons can be created based on a reference video. }


\rev{\paragraph{Runtime speed} 
Generating a video from a $15$ second input audio and a $2$ minute reference video takes about $43$ seconds in total. Here is the breakdown: (a) $8$ seconds are used for audio-driven search to find graph paths in the video motion graph, (b) $35$ seconds are used for synthesizing all transitions. Specifically, for a $15$ second input audio, there are maximum of $4$ synthetic transitions in our examples, with $8$ blended frames created per transition. For blending, obtaining the initial mesh flow from human fitting takes $1$ second, then synthesizing each blended frame takes $0.1$ seconds measured on a single Tesla V100 GPU. }

\paragraph{Personal data / human subjects.} \rev{The Personal story dataset contains 7 videos with 6 different speakers (5 male, 1 female). The number of frames ranges from 4465 to 19176 (148 to 639 seconds).} We collected it under the permission from each speaker to include frames, clips and full video in the paper submission. We also used the TED-talks dataset from the previous work \cite{siarohin2021motion}. The perceptual user study is collected with the approval of IRB.

\end{document}